%% file: main.tex
\pdfoutput=1
\documentclass[11pt]{article}

\usepackage[final]{acl}

\usepackage{times}
\usepackage{latexsym}

\usepackage[T1]{fontenc}
\usepackage[utf8]{inputenc}

\usepackage{microtype}

\usepackage{nccmath}
\usepackage{amsmath}
\usepackage{amsfonts}
\usepackage{amssymb}
\usepackage{graphicx}
\usepackage{float}
\usepackage{subfigure} 
\usepackage[ruled]{algorithm2e}
\usepackage{algpseudocode}
\usepackage{multirow}
\usepackage{array}
\usepackage{siunitx}
\usepackage{booktabs}
\usepackage{hyperref}
\usepackage{enumitem}
\usepackage{balance}

\usepackage{appendix}
\usepackage{titletoc}

\newcommand{\hide}[1]{} %hide

\newcommand{\vpara}[1]{\vspace{1.5ex}\noindent\textbf{#1}}

\newcommand{\model}{SP$^3$}
\newcommand{\smodel}{SP$^3$ }
\title{\model: Enhancing Structured Pruning via PCA Projection}

\author{
    Yuxuan Hu\textsuperscript{\rm 1,\rm 2},\ Jing Zhang\textsuperscript{\rm 1,\rm 3}\thanks{\ \ Corresponding author.},\ Zhe Zhao\textsuperscript{\rm 4},\ Chen Zhao\textsuperscript{\rm 1,\rm 2}, \\
    \  {\bf Xiaodong Chen}\textsuperscript{\rm 5},\ {\bf Cuiping Li}\textsuperscript{\rm 1,\rm 3},\ {\bf Hong Chen}\textsuperscript{\rm 1,\rm 3}\thanks{\ \{huyuxuan1999,zhang-jing\}@ruc.edu.cn, \\ nlpzhezhao@tencent.com, zhaochen100@ruc.edu.cn, chenxiaodong6677@163.com, \{licuiping,chong\}@ruc.edu.cn} \\
    \textsuperscript{\rm 1}School of Information, Renmin University of China, Beijing, China \\
    \textsuperscript{\rm 2}Key Laboratory of Data Engineering and Knowledge Engineering, MOE, China \\
    \textsuperscript{\rm 3}Engineering Research Center of Database and Business Intelligence, MOE, China \\
    \textsuperscript{\rm 4}Tencent AI Lab, Tencent, Beijing, China \\
    \textsuperscript{\rm 5}School of Computer Science and Technology, Xi'an Jiaotong University, Xi'an, China \\
}

\begin{document}
\maketitle

\input{contents/abstract}
\input{contents/introduction}

\input{contents/background}
\input{contents/method}

\input{contents/experiments}

\input{contents/relatedwork}
\input{contents/conclusion}

\section*{Limitations}

Our proposed \smodel introduces extra weight matrices in the residual parts when merging the inserted compactor matrices with the weight matrices. When the model is compressed to 5M, these extra parameters account for more than half of the model. This predominance hinders further compression. In future research, we aim to explore strategies to eliminate these extraneous parameters. Meanwhile, the efficient acceleration of the model mainly relies on coarse-grained pruning (head-level, layer-level), while our approach uses more fine-grained pruning, and how to balance the fine-grained and coarse-grained pruning to achieve higher acceleration also needs further exploration.

When applying to large language models, integrating compactors during training in our methodology may lead to increased memory consumption, necessitating efforts to reduce memory overhead. Additionally, experimental results indicate that our tuning-based pruning technique may face challenges with overfitting, particularly when tuning using a limited SFT dataset. Therefore, investigating a tuning-free pruning technique is warranted.

\section*{Ethical Considerations}

\vpara{Intellectual Property.} The datasets we use include oasst\_top1\_2023-08-25, MNLI, QNLI, QQP, SST-2, MRPC, STS-B, RTE, SQuAD, ARC-e, ARC-c, BoolQ, HellaSwag, openbookQA, PIQA, and WinoGrande. The models we use include $\rm BERT_{base}$, $\rm OPT_{125m}$, and TinyLlama. These are publicly accessible and well-established resources aimed at facilitating diverse AI and NLP research endeavors. 

\vpara{Data annotation.} We utilize the annotations provided by existing datasets, thereby eliminating the need for manual annotation in our study.

\vpara{Intended Use.} $\rm SP^3$ is a model pruning technique designed to compress the parameters of a transformer model while preserving performance, effectively reducing the computational resources needed for deploying the model.

\vpara{Misuse risks.} $\rm SP^3$ is a pruning method for transformer models and incorrect use of $\rm SP^3$ might degrade the performance of some applications.

\vpara{Misuse Control.} We intend to make our approach available to the open-source community, enabling users to gain a deeper understanding of our methodology and mitigate the risk of misuse.

\section*{Acknowledgments}

This work is supported by the National Key Research \& Develop Plan (2023YFF0725100) and the National Natural Science Foundation of China (62322214, U23A20299, 62076245, 62072460, 62172424, 62276270). This work is supported by Public Computing Cloud, Renmin University of China.  

\bibliography{anthology,custom}
\bibliographystyle{acl_natbib}

% \newpage

\appendix
% \startcontents[appendix]
% \printcontents[appendix]{ }{1}{\setcounter{tocdepth}{2}\vspace{5pt}}
\input{appendix/MHA_pca_projection}
\input{appendix/pseudo_code}
\input{appendix/sparsity}
\input{appendix/experiment}
\input{appendix/more_result}
\input{appendix/SP3_for_LLM}
\input{appendix/more_ablation}
\input{appendix/structures}
\input{appendix/low_rank}
\input{appendix/figures}

\end{document}

%% file: contents/abstract.tex
\begin{abstract}

% Weight-Inherited Distillation (WID) is an effective distillation method that inherits the weights from the teacher's model, thus achieving better results than traditional distillation methods. However, the identity matrix initialization used in WID leads to slow model convergence. In this work, we propose an improved WID method named \smodel that replaces the identity matrix initialization with a specialized data-aware initialization. Concurrently, we refine the structural design of WID, enhancing its adaptability and flexibility in selecting the compressed model architecture. 

Structured pruning is a widely used technique for reducing the size of pre-trained language models (PLMs), but current methods often overlook the potential of compressing the hidden dimension (\(d\)) in PLMs, a dimension critical to model size and efficiency. This paper introduces a novel structured pruning approach, \underline{S}tructured \underline{P}runing with \underline{P}CA \underline{P}rojection (\model), targeting the effective reduction of \(d\) by projecting features into a space defined by principal components before masking. 
Extensive experiments on benchmarks (GLUE and SQuAD) show that \smodel can reduce \(d\) by 70\%, compress 94\% of the $\rm BERT_{base}$ model, and maintain over 96\% accuracy and outperform other methods that compress \(d\) by 6\% in accuracy at the same compression ratio. \smodel has also proven effective with other models, including OPT and Llama.
Our data and code are available at \href{https://github.com/hyx1999/SP3}{ours repo}.

\end{abstract}

%% file: contents/introduction.tex
\begin{figure}[t] %H为当前位置，!htb为忽略美学标准，htbp为浮动图形
\centering %图片居中
\includegraphics[width=0.4\textwidth]{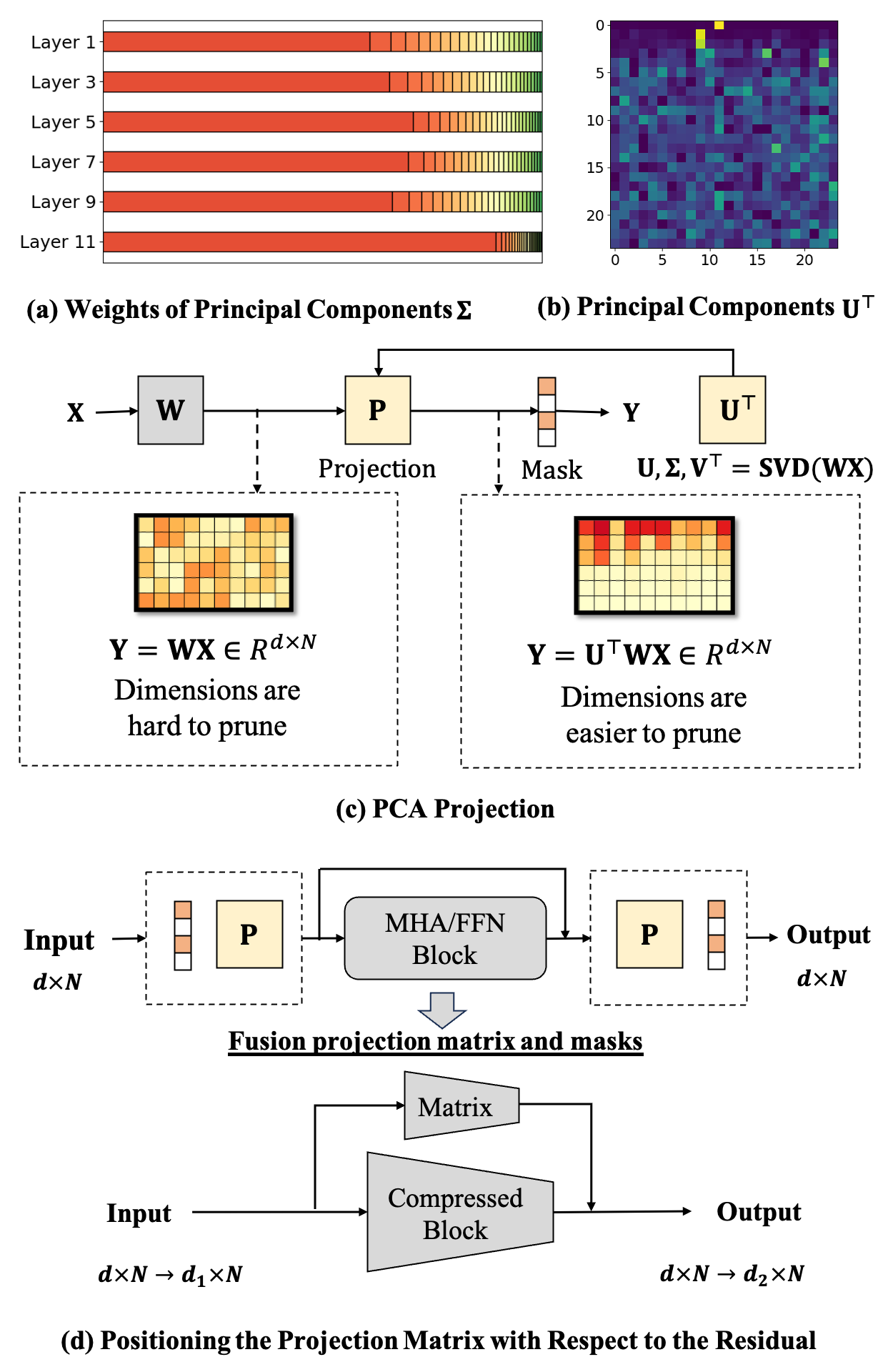} %插入图片，[]中设置图片大小，{}中是图片文件名
\caption{
(a) Depiction of principal component weights within $\Sigma$. The length of each rectangle responds to the percentage of that principal component's weight among all principal components;
(b) Illustration of the principal component matrix. Each row of the matrix represents a principal component. Brighter colors indicate a higher correlation between the feature dimension and the principal component under investigation.
% (b) Illustration of principal component weights in $U$ per layer. Each row represents a principal component, each column represents a dimension, and the colors reflect the correlation between the principal component and the dimension;
(c) Projection of features onto principal components for easier dimension pruning;
(d) Positioning the projection matrix outside the residual.
In the figure, {\small$d,d_1,d_2$} denote the dimensions of the features and {\small$N$} denotes the number of tokens.
} %最终文档中希望显示的图片标题
\label{Fig.intro} %用于文内引用的标签
\end{figure}

\section{Introduction}
Pre-trained language models (PLMs) like BERT \cite{BERT.devlin2018}, GPT-2 \cite{GPT2.radford2019}, ChatGPT \cite{instruct-gpt.ouyang2022}, and Llama \cite{llama.touvron2023}, leveraging transformer architectures, have gained prominence in Natural Language Processing due to their exceptional performance. However, their substantial storage requirements and long inference times present practical challenges, prompting the need for efforts to optimize their size and computational speed.

Among various model compression techniques, distillation~\cite{Distillbert.VictorSanh} and, pruning~\cite{l0pruning.louizos2018}, particularly structural pruning~\cite{cofi.xia2022}, are prominent. Distillation is adaptable, allowing for the design of compact student models while transferring knowledge from larger teacher models. Nevertheless, it might not fully leverage the teacher model's parameters, often requiring the student model to undergo extensive training on large datasets. In contrast, structured pruning directly trims the original model, thereby utilizing its parameters more efficiently and lessening the dependency on large datasets.

The transformer architecture in PLMs is defined by weight matrices controlled by hidden dimension $d$, attention head size $d_h$ in multi-head attention layers, and filter dimension $d_f$ in feed-forward network layers. Focused efforts in structured pruning have aimed at downsizing $d_h$ and $d_f$~\cite{HeadPruning.michel2019, Fast.kwon2022, HeadPruning.voita2019, Structured.wang2020}. Yet, strategies for pruning $d$ are underexplored, with limited compression success. For instance, CoFi~\cite{cofi.xia2022} achieves a notable 95\% sparsity and maintains over 90\% accuracy but only marginally reduces $d$ from 768 to 760. 
We posit that the hidden dimension \(d\) holds significant potential for compression. Recent research demonstrates that high-dimensional features produced by models can be projected into a subspace, leveraging their inherent low-rank properties~\cite{Drone.chen2021}. Our observations further reveal that the sizes of these subspaces can differ among various layers. This variability underscores the necessity of compressing \(d\) within each individual layer, as the distribution of principal components' weights changes across layers, a phenomenon depicted in Fig. \ref{Fig.intro} (a).
Unfortunately, presently, distillation remains the most effective means for compressing $d$. This paper aims to explore a potent structured pruning method for efficiently reducing $d$ in PLMs.

% We argue that compressing \(d\) could significantly impact model compression performance because this compression can be viewed as a projection of features from the original space to a lower-dimensional subspace.

We analyze and conjecture that existing pruning methods rarely compress \(d\) effectively for two main reasons. First, conventional mask-based pruning methods directly reduce dimensions in the original feature space, which doesn't effectively eliminate unimportant dimensions. We interpret the pruning of \(d\) from the perspective of features' principal components. Drawing inspiration from PCA theory \cite{PCA}, effective compression should preserve the principal components of the features. Considering that most dimensions in the original feature space are linked to these principal components, as indicated by the evenly distributed values across each row that represents a principal component in Fig.~\ref{Fig.intro}(b), traditional mask-based pruning in this space struggles to preserve the principal components' integrity. Secondly, the residual structure in each block, including both the multi-head attention (MHA) and the feed-forward network (FFN), requires the input and output dimensions of each block to be identical, hindering the pruning of different \(d\) across various layers.

To overcome these limitations, we propose an enhanced \underline{S}tructured \underline{P}runing approach using \underline{P}CA \underline{P}rojection (\model) to reduce the hidden dimension. This method involves projecting the original features into a new space, where principal components act as the basis vectors. In this way, mask-based structured pruning becomes more effective in eliminating less important dimensions while retaining the principal components. For the projection, we select several tokens from the dataset designated for pruning and process them through the transformer model to acquire their feature matrix.
Subsequently, we apply SVD to this matrix to obtain the projection matrix, which is then integrated before the mask matrix in the conventional pruning methodology. This process enables the projection of the feature matrix into a space defined by the principal components (as depicted in Fig.\ref{Fig.intro} (c)).
Additionally, to enable different \(d\) in different layers, we introduce additional linear transformations in the residuals, resulting in layer-adapted structured pruning (as shown in Fig.\ref{Fig.intro} (d)). 

The primary contributions of this work are:
\begin{itemize}[leftmargin=*]
\item 
 We take a fresh perspective on structured pruning by reducing the hidden dimension of the transformer model, which has been rarely explored before. 

\item Our \smodel method enhances compression efficacy by projecting features into space formed by principal components before masking, which helps preserve the principal components of features. Furthermore, it integrates linear transformations in residual connections, enabling variable hidden dimensions across different layers for a more customized structured pruning approach.

\item Extensive testing on the GLUE and SQuAD benchmarks reveals that \smodel can effectively reduce the hidden dimension by 70\%, compressing 94\% of the $\rm BERT_{base}$  model while still maintaining an accuracy of over 96\%. When compared to other methods that also compress the hidden dimension (\(d\)), \smodel demonstrates a 6\% improvement in accuracy at the same compression ratio.
\end{itemize}

%% file: contents/background.tex
\begin{figure*}[t]
    \centering
    \includegraphics[width=\textwidth]{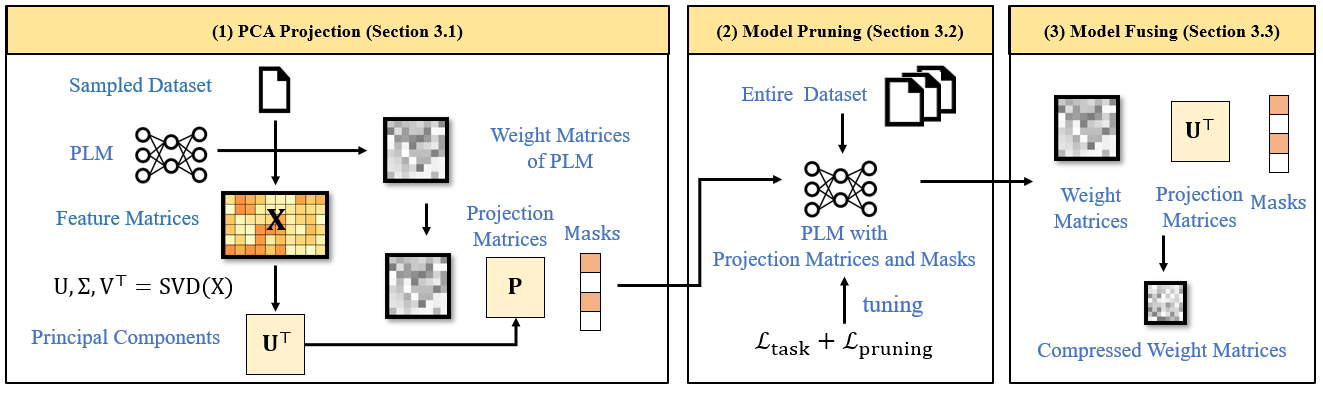}
    \caption{\label{Fig.workflow} Illustration of the workflow of \smodel.}
\end{figure*}

\begin{figure}[t]
    \centering
    \includegraphics[width=0.5\textwidth]{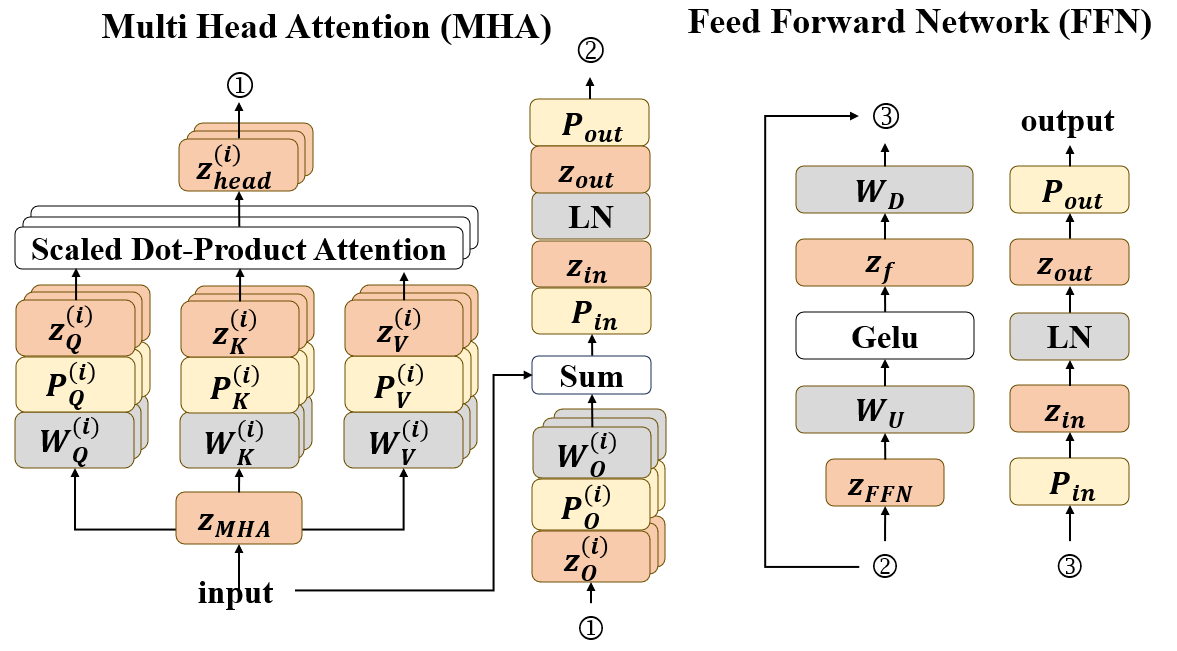}
    \caption{Illustration of the \smodel architecture, in which the gray rectangles represent the weight matrices, the yellow rectangles signify the projection matrices, and the red rectangles indicate the masks.}
    \label{Fig.model-structure}
\end{figure}

\section{Background}
\label{sec:background}

\subsection{Transformer Architecture}

A typical transformer comprises a stack of transformer layers, each containing an MHA block and an FFN block. We take BERT \cite{BERT.devlin2018} as an example to describe this architecture.

\vpara{Multi Head Attention (MHA)}. The MHA block takes $x \in \mathbb{R}^{d \times N}$ as input and consists of $H$ attention heads that facilitate interactions between tokens, with a LayerNorm \cite{Layernorm.Jimmy2016} step. 

% \label{Background.MHA.x}

\begin{small}
\begin{align}
\begin{split}
    & x_{\rm M} = {\rm LN}(x + {\rm MHA}(x)), \\
    & {\rm MHA}(x) = \sum_{i=1}^{H} {\rm Att}^{(i)}(x), \\
    & {\rm Att}^{(i)} (x) = W_O^{(i) {\top}} W_V^{(i)} x \cdot {\rm Softmax}( \frac{(W_K^{(i)} x)^{\top} (W_Q^{(i)} x)}{\sqrt{d_h}} ), 
\end{split}
\end{align}
\end{small}

\noindent where {\small $W_Q^{(i)}, W_K^{(i)} \in \mathbb{R}^{d_{\rm QK} \times d}$, $ W_V^{(i)}, W_O^{(i)} \in \mathbb{R}^{d_{\rm VO} \times d} $ } denote the query, key, value, and output matrices, respectively, $d$ denotes the hidden dimension, $d_h = d / H$ denotes the head size, $N$ denotes the sequence length, and $d_{\rm QK}$ and $d_{\rm VO}$ denote the intermediate dimensions.

% , denoting the query, key, value, and output matrices, respectively

\vpara{Feed Forward Network (FFN)}. The FFN block takes $x_M$ as input and generates $x_F$ as output:

% \label{Background.FFN.x}
\begin{small}
\begin{align}
\begin{split}
    & x_{\rm F} = {\rm LN}(x_{\rm M} + {\rm FFN}( x_{\rm M} )), \\
    & {\rm FFN}(x_{\rm M}) = W_{\rm D}^{\top} {\rm gelu} (W_{\rm U} x_{\rm M}),
\end{split}
\end{align}
\end{small}

\noindent where $\rm gelu$ is the activation function, and $W_U, W_D \in \mathbb{R}^{d_f \times d}$ are parameters of an FFN block, $d_f$ indicates the intermediate dimension.

% two weight matrices

\subsection{Mask-based Structured Pruning}
Mask-based structured pruning approaches eliminate parameters by integrating additional mask variables $z$ into the original model's parameters, as follows:

\begin{small}
\begin{align}
\begin{split}
    & x_{\rm M} = {\rm diag}(z_{\rm out}) {\rm LN}({\rm diag}(z_{\rm in})(x + z_{\rm MHA} {\rm MHA}(x))), \\
    & {\rm MHA}(x) = \sum_{i=1}^{H} z_{\rm head}^{(i)} {\rm Att}^{(i)}(x), \\
    & x_{\rm F} = {\rm diag}(z_{\rm out}) {\rm LN}({\rm diag}(z_{\rm in})(x_{\rm M} + z_{\rm FFN} {\rm FFN}( x_{\rm M} ))), \\
    & {\rm FFN}(x_{\rm M}) = W_{\rm D}^{\top} {\rm diag}(z_{\rm f}) {\rm gelu} (W_{\rm U} x_{\rm M}), 
    \label{Eq.masked_base_pruning}
\end{split}
\end{align}
\end{small}

\noindent where {\small $z_{\rm MHA},z_{\rm FFN} \in \{0, 1\}, z_{\rm head} \in \{0, 1\}^{H}, z_{\rm f} \in \{0, 1\}^{d_f}, z_{\rm in}, z_{\rm out}\in \{0, 1\}^{d}$ } are used to remove entire $\rm MHA$ block, $\rm FFN$ block, attention head, intermediate dimensions of FFN block, and hidden dimensions, respectively. Then, the pruning process can be modeled as an optimization problem with sparsity constraints as follows,

% remove entire $\rm MHA$ block, $\rm FFN$ block, attention head, intermediate dimensions of FFN block, and hidden dimensions respectively

% \mathop{\arg\min}_{\theta, z}
\begin{small}
\begin{align}
\begin{split}
    & \min \mathcal{L}(\theta) \\
    & \text{s.t.} \quad {\rm sparsity}(z) \geq t, 
\end{split}
\end{align}
\end{small}

\noindent where $\theta$ denotes the parameters of the model, and $t$ denotes the pre-defined target sparsity.

% WID is a distillation method that, in contrast to conventional distillation methods, inherits the teacher model's parameters and compresses them.

% Figure \ref{Fig.1} (a) illustrates the core concept of WID's model compression. Initially, WID introduces compactor matrices $W_L \in \mathbb{R}^{d \times d}$ and $W_R \in \mathbb{R}^{d \times d}$, along with associated masks $M_L \in \mathbb{R}^{d}$ and $M_R \in \mathbb{R}^{d}$, positioned on either side of the transformer block. The compactor matrices are set to the identity matrix at the start, while the masks begin as vectors filled with ones. After this setup, the model undergoes training using a large dataset, and compactor matrices are progressively pruned based on the masks. The final step involves fusing the transformer block with the compactors to produce a compressed transformer block. It can be seen that the input and output of the transformer block are compressed from $d$ to $d_0$ dimension.

%% file: contents/method.tex
\section{\model}

%The central principle of \smodel involves projecting features into a space defined by their principal components, followed by pruning their dimensions within that space. This requires the integration of PCA projections into the model's computation process. 
%Additionally, it's crucial that these projections do not alter the original model's output and can be smoothly integrated with the original model's parameters.

\vpara{Workflow of \model.} The \smodel we propose unfolds in three phases, as depicted in Fig. \ref{Fig.workflow}. Initially, PCA projection is conducted by selecting a subset of the training data, processing it through the transformer model to yield the feature matrix \( X \) for each block per layer, and then determining each \( X \)'s principal components \( U \) via SVD decomposition. Following this, a projection matrix, initialized with these principal components, is integrated into the model's computational framework, coupled with a mask for structured pruning, detailed in Section~\ref{sec:pca_projection}. Subsequently, the model undergoes pruning based on the masks tailored to a specific task and pruning objective, as outlined in Section~\ref{section:pruning}. The final stage involves merging these masks and projection matrices with the original model parameters, thereby creating the compressed model, which is further elaborated in Section~\ref{section:fusing}.

\subsection{PCA Projection}
\label{sec:pca_projection}

Current structured pruning methods typically compress PLMs by directly applying a mask to their hidden dimensions. However, this approach overlooks the relationship between each feature dimension and its corresponding principal component. Principal component analysis of these features reveals that a few principal components hold the majority of the weight in every layer (as indicated in Fig.~\ref{Fig.intro}(a)). Additionally, these principal components are connected to a large number of dimensions, as evidenced by the uniformly distributed values across each row representing a principal component (see Fig.~\ref{Fig.intro}(b)). Consequently, directly pruning dimensions can adversely affect several principal components, thereby complicating the compression of hidden dimensions.

To address this, we incorporate a projection matrix prior to each masking matrix in the computational process of PLMs, referred to as PCA Projection. This matrix is designed to project features into a space determined by their principal components. Subsequently, these principal components can be pruned directly using masks. It's crucial to ensure that these added projection matrices do not alter the output of the PLM and can be seamlessly integrated with the model's parameters. In the following section, we detail the methodology for injecting these projection matrices to meet both of these conditions.

\vpara{Hidden Dimension PCA Projection.} To decrease the hidden dimension (\(d\)) between layers, our initial step involves selecting calibration data from the training dataset. We then derive the corresponding feature matrix \(X \in \mathbb{R}^{d \times T}\) from either the MHA block or the FFN block, along with their respective residuals. Here, \(T\) represents the number of sampled tokens. We choose \(T\) to be marginally greater than \(d\) to guarantee that the number of tokens exceeds the number of feature dimension. This feature matrix $X$ acts as the input for the {\small$\rm LayerNorm$} layer. Subsequently, we apply SVD to extract the principal components of the feature. Each column of the matrix \( U \) represents a principal component, obtained from the SVD of the matrix \( X \), as described by \( U, \Sigma, V^{\top} = \text{SVD}(X - \overline{X}) \), where \( \overline{X} \) represents the matrix obtained by broadcasting the average values of each column.

% (i.e., Eq.~\ref{Background.MHA.2}) , (i.e., Eq.~\ref{Background.FFN.2})

Next, we introduce the matrix $U$ both before and after each {\small$\rm LayerNorm$} layer as follows. Initially, we recognize:

\begin{small}
\begin{align}
\begin{split}
{\rm LN}(X) & = {\rm diag}(\gamma) {\rm norm}(X - \overline{X}) \\
            & = {\rm diag}(\gamma) {\rm norm}(U U^T (X - \overline{X})),
\end{split}
\end{align}    
\end{small}

\noindent where {\small${\rm LN}$} signifies {\small$\rm LayerNorm$} module (the bias term is omitted for simplicity), {\small$\rm norm$} represents the normalization function defined as {\small${\rm norm}(x) = x / \Vert x \Vert * d$ }, and $UU^T=I$ because columns of $U$ are orthogonal. Given the following formulas, 

\begin{small}
\begin{align}
\begin{split}
{\rm norm}(U x) & = U x / \Vert U x \Vert * d \\
& = U x / \sqrt{(x^{\top} U^{\top} U x)} * d \\
& = U x / \sqrt{(x^{\top} x)} * d \\
& = U {\rm norm}(x),
\end{split}
\end{align}    
\end{small}

\begin{small}
\begin{align}
\begin{split}
X - \overline{X} & = (I - \frac{1}{d}11^{\top})X \\
& = R X,
\end{split}
\end{align}    
\end{small}

\noindent where we define {\small $R =  (I - \frac{1}{d}11^{\top})$} using \( I \) to represent the identity matrix and \( 1 \) to denote a vector filled with ones. Then, we have,

\begin{small}
\begin{align}
{\rm LN}(X) = {\rm diag}(\gamma) U {\rm norm}(U^{\top} R X).
\end{align}
\end{small}

\noindent Based on this, we assign projection matrices {\small$P_{\rm in} = U^{\top} R$} and {\small$P_{\rm out} = {\rm diag}(\gamma) U$}, respectively. 

The computation process after introducing the hidden dimension PCA Projection is as follows,

\begin{small}
\begin{align}
\begin{split}
    & x_{\rm M} =  P_{\rm out}^{\rm M} {\rm diag}(z_{\rm out}^{\rm M}) {\rm LN}({\rm diag}(z_{\rm in}^{\rm M}) P_{\rm in}^{\rm M} (x + z_{\rm MHA}{\rm MHA}(x))),  \\
    & x_{\rm F} = P_{\rm out}^{\rm F} {\rm diag}(z_{\rm out}^{\rm F}) {\rm LN}( {\rm diag}(z_{\rm in}^{\rm F}) P_{\rm in}^{\rm F} ( x_{\rm M} + z_{\rm FFN}{\rm FFN}( x_{\rm M} ))),
    \label{Eq.hidden_dimension_pca_projection}
\end{split}
\end{align}
\end{small}

\noindent where the superscripts $^{M}$ and $^{F}$ denote the projection and mask matrices within the MHA and FFN blocks, respectively.

\vpara{MHA Intermediate Dimension PCA Projection.} While our focus is primarily on compressing the hidden dimension, our method is also applicable to compressing the intermediate dimension of the MHA block. We begin by collecting the feature matrix  {\small$X \in R^{d \times N}$} from the output preceding the MHA block. We aim to project the features {\small $W_Q^{(i)} X$, $W_K^{(i)} X$} and {\small $W_V^{(i)} X$} using their principal components {\small $U_Q^{(i)}, U_K^{(i)}$}, and {\small $U_V^{(i)}$} respectively. They should satisfy the following equations:

\begin{small}
\begin{align}
\begin{split}
W_V^{(i)} X & = U_V^{(i)\top} U_V^{(i)} W_V^{(i)} X, \\
X^{\top} W_Q^{(i)\top} W_K^{(i)} X & = X^{\top} W_Q^{(i)\top} U_Q^{(i)\top} U_K^{(i)} W_K^{(i)} X, \label{Eq.PCA.U_qkv}
\end{split}
\end{align}
\end{small}

\noindent where {\small$U_V^{(i)}$} in the equation can be solved directly by SVD, i.e., {\small $U_V^{(i)} = U$, where $U, \Sigma, V^{\top} = {\rm SVD}(W_V X)$}. The solution of $U_Q^{(i)}, U_K^{(i)}$ in Eq. \ref{Eq.PCA.U_qkv} is a more complex, and we follow the method in DRONE \cite{Drone.chen2021} to compute {\small$U_Q^{(i)}, U_K^{(i)}$}:

\begin{small}
\begin{align}
\begin{split}
U_Q^{(i)} & = \Sigma_{Z}^{(i)\frac{1}{2}} U_{Z}^{(i)\top} \Sigma_1^{(i)-1} U_1^{(i)\top}, \\
U_K^{(i)} & =  \Sigma_{Z}^{(i)\frac{1}{2}} V_{Z}^{(i)} \Sigma_2^{(i)-1} U_2^{(i)\top}, 
\label{Eq.MHA_PCA_projection}
\end{split}
\end{align}    
\end{small}

\noindent where

\begin{small}
\begin{align}
\begin{split}
    U_1^{(i)}, \Sigma_1^{(i)}, V_1^{(i)\top} & = {\rm SVD}(W_Q^{(i)} X), \\
    U_2^{(i)}, \Sigma_2^{(i)}, V_2^{(i)\top} & = {\rm SVD}(W_K^{(i)} X), \\
    Z^{(i)} & = \Sigma_1^{(i)\top} U_1^{(i)\top} U_2^{(i)} \Sigma_2^{(i)}, \\
    U_{Z}^{(i)}, \Sigma_{Z}^{(i)}, V_{Z}^{(i)\top} & = {\rm SVD}(Z^{(i)}).
\end{split}
\end{align}    
\end{small}

\noindent We then assign {\small$P_Q^{(i)} = U_Q^{(i)}$, $P_K^{(i)} = U_K^{(i)}$, $P_V^{(i)} = U_V^{(i)\top}$}, and {\small$P_O^{(i)} = U_V^{(i)}$} as our target projection matrices. For the proof of Eq. \ref{Eq.MHA_PCA_projection}, refer to Appendix \ref{Appendix.MHA_PCA_Projection}.

The computation process incorporating  MHA intermediate dimension PCA Projection is as follows,

\begin{small}
\begin{align}
\begin{split}
    & {\rm Att}^{(i)} (x) = W_O^{(i) \top} P_O^{(i) \top} {\rm diag}(z_O^{(i)})^{\top} {\rm diag}(z_V^{(i)}) P_V^{(i)} W_V^{(i)} x \cdot \\
    & {\rm Softmax}( \frac{(W_K^{(i)} x)^{\top} P_K^{(i) \top} {\rm diag}(z_K^{(i)})^{\top} {\rm diag}(z_Q^{(i)}) P_Q^{(i)} (W_Q^{(i)} x)}{\sqrt{d_h}}). 
    \label{Eq.MHA_intermediate_pca_projection}
\end{split}
\end{align}
\end{small}

\vpara{Model Structure.} We integrate the aforementioned PCA projections across various dimensions and present the final structure of our model in Fig. \ref{Fig.model-structure}. Unlike mask-based pruning methods such as CoFi, which employ a shared mask in each block, our approach utilizes distinct masks for different blocks and incorporates specific projection matrices for each block (as contrasted in Eq. \ref{Eq.masked_base_pruning} and Eq. \ref{Eq.hidden_dimension_pca_projection}). This strategy enables us to independently compress the hidden dimension of each block. Additionally, we insert projection matrices in the MHA block to compress its intermediate dimension (as detailed in Eq. \ref{Eq.MHA_intermediate_pca_projection}). The transformer’s residual structure necessitates the same hidden dimension across different blocks. By introducing a linear transformation in the residuals, we effectively circumvent this constraint (as illustrated in Figure~\ref{Fig.intro}(d)).
Readers interested in the algorithmic details can find the pseudo-code for the PCA Projection in the Appendix~\ref{Appendix.code}.

\vpara{Complexity.} The computational complexity of PCA Projection primarily hinges on three factors: the number of layers \( n \), the hidden dimension \( d \), and the number of sampled tokens \( T \), resulting in an approximate complexity of \( O(n*d^2*T) \). In practice, for a model with 100M parameters, the PCA Projection process takes around 5 minutes. For a larger model with 7B parameters, we estimate that the PCA Projection would require roughly 1 hour, significantly less than the time required for model pruning.

\subsection{Model Pruning}
\label{section:pruning}

\vpara{Objective Function}. Pruning is governed by the masks incorporated into the model, and the model's sparsity is determined based on these masks. Throughout the training process, all mask variables are treated as continuous real numbers. Upon completion of the training, those mask variables falling below a certain threshold — set in accordance with the desired sparsity level — are converted to 0, thereby establishing the final pruned structure. We use pruning loss \cite{Structured.wang2020} that forces the expected sparsity of the model to be close to the desired sparsity:

\begin{small}
\begin{align}
    \mathcal{L}_{\rm pruning} = \lambda_1 \cdot (\hat{s} - t) + \lambda_2 \cdot (\hat{s} - t)^2,
\end{align}
\end{small}

\noindent where $\hat{s}$ is the expected sparsity, $t$ is the target sparsity, and $\lambda_1, \lambda_2$ are two Lagrange multipliers. Please refer to Appendix \ref{Appendix.Sparsity} for more detail of $\hat{s}$. Additionally, we incorporate the same task-specific loss, {\small$\mathcal{L}_{task}$}, as used in CoFi \cite{cofi.xia2022}, applied to the task-oriented dataset during the pruning process to maintain model performance on specified tasks.

\vpara{Multi-stage Pruning}. In \model, we implement masks at three hierarchical levels: dimension-level (\( z_{\rm in} \), \( z_{\rm out} \), \( z_{\rm f} \), \( z_{\rm Q} \), \( z_{\rm K} \), \( z_{\rm V} \), \( z_{\rm O} \)), head-level (\( z_{\rm head} \)), and layer-level (\( z_{\rm MHA} \), \( z_{\rm FFN} \)). Our findings reveal that during training, the model prioritizes head-level and layer-level masks for compression, while dimension-level masks are sometimes less emphasized. Consequently, we have divided the training process into two distinct stages. In the first stage, we exclusively employ dimension-level masks, and then, in the subsequent stage, we add head-level and layer-level masks. Empirically, the initial stage spans 5 epochs, followed by the second stage, which lasts for 15 epochs.

\subsection{Model Fusing}
\label{section:fusing}

After training, we first prune projection matrices, attention heads, MHA blocks, and FFN blocks based on different masks. After that, we merge the projection matrices with weight matrices. Suppose function $\rm Pr$ denotes pruning all zero rows and columns of the matrix.

\vpara{MHA Layer Fusing.} For the MHA layer, we have

\begin{small}
\begin{align}
\begin{split}
    \hat{W}_Q^{(i)} & = {\rm Pr}({\rm diag}(z_{\rm Q}) P_Q W_Q^{(i)} P_{\rm out}^{\rm M} {\rm diag}(z_{\rm out}^{\rm M})), \\
    \hat{W}_K^{(i)} & = {\rm Pr}({\rm diag}(z_{\rm K}) P_K W_K^{(i)} P_{\rm out}^{\rm M} {\rm diag}(z_{\rm out}^{\rm M})), \\
    \hat{W}_V^{(i)} & = {\rm Pr}({\rm diag}(z_{\rm V}) P_V W_V^{(i)} P_{\rm out}^{\rm M} {\rm diag}(z_{\rm out}^{\rm M})), \\
    \hat{W}_O^{(i)\top} & = {\rm Pr}({\rm diag}(z_{\rm in}^{\rm F}) P_{\rm in}^{\rm F} W_O^{(i)\top} P_O {\rm diag}(z_{\rm O})), 
\end{split}
\end{align}
\end{small}

\noindent where {\small$\hat{W}_Q^{(i)}, \hat{W}_K^{(i)}, \hat{W}_V^{(i)}$, and $ \hat{W}_O^{(i)}$} denotes the compressed weight matrices of the MHA block.

\vpara{FFN Layer Fusing}. For the FFN layer, we have

\begin{small}
\begin{align}
\begin{split}
    \hat{W}_U & = {\rm Pr}({\rm diag}(z_f) W_U P_{\rm out}^{\rm M} {\rm diag}(z_{\rm out}^{\rm M})), \\
    \hat{W}_D^\top & = {\rm Pr}({\rm diag}(z_{\rm in}^{\rm F}) P_{\rm in}^{\rm F} W_D^\top {\rm diag}(z_f)), 
\end{split}
\end{align}
\end{small}

\noindent where {\small$\hat{W}_U$} and {\small$\hat{W}_D$} denotes the compressed weight matrices of the FFN block.

\vpara{Residual Modification.} Since we adaptively learn different hidden dimensions in different layers, we need to add weight matrices to the residual. For example, for $i$-th block of Transformer, we have

\begin{small}
\begin{align}
    x^{(i)} = & {\rm LN}(W_{\rm R}^{(i)} x + {\rm Block}^{(i)}(x^{(i-1)}))), 
\end{align}
\end{small}

\noindent where $W_{\rm R}$ is a matrix added to the residual, which is determined by multiplying the pruned projection matrices before and after the block (refer to Figure~\ref{Fig.intro} (d)), i.e.,

\begin{small}
\begin{align}
    W_{\rm R}^{(i)} = {\rm Pr}({\rm diag}(z_{\rm in}^{(i)}) P_{\rm in}^{(i)} P_{\rm out}^{(i-1)} {\rm diag}(z_{\rm out}^{(i-1)})).
\end{align}
\end{small}

%% file: contents/experiments.tex
\section{Experiments}

\subsection{Setup}

% We exclude CoLA due to their unstable behaviors, and we cannot reproduce some baseline results based on our device on the CoLA dataset.

\vpara{Datasets.} We evaluate our approach on GLUE \cite{GLUE.wang2018} tasks and SQuAD \cite{SQuAD.2016} v1.1. GLUE tasks include SST-2 \cite{sst2.socher2013}, MNLI \cite{mnli.kim2019}, QQP \cite{qqp.wang2017}, QNLI, MRPC \cite{mrpc.dolan2005}, STS-B, and RTE. Each comparison model is evaluated on the test set of these datasets. For the GLUE benchmark, accuracy metrics are used for MNLI, QQP, QNLI, SST2, MRPC, and RTE tasks, while Spearman's correlation is reported for STS-B. In the case of the SQuAD benchmark, we present the F1 score. For comprehensive details about these datasets, please refer to Appendix \ref{appendix.dataset}.

\vpara{Baselines.} We evaluate the proposed \smodel against established distillation methods like TinyBERT \cite{TinyBERT.jiao2020}, WID \cite{WID.wu2023}, and the pruning approach CoFi \cite{cofi.xia2022}.

% structured pruning

\vpara{Settings.} Our primary experiments focus on compressing the $\rm BERT_{base}$ model. \smodel and CoFi perform model pruning using the previously mentioned training datasets. In contrast, WID and TinyBERT additionally use more general pre-training data. For a fairer comparison, we have excluded the data augmentation component from TinyBERT in our experiments. To generate calibration data for the SVD decomposition in \model, we randomly choose 512 samples from each dataset and concatenate them, and then from them, we randomly select 4,096 tokens as our calibration dataset. For detailed information on our experimental methods, please see the Appendix~\ref{appendix.setup}.

\subsection{Main Results}

% -----------------------------------------------------------------------------------------
\begin{figure*}[t] %H为当前位置，!htb为忽略美学标准，htbp为浮动图形

\subfigure[Hidden dimensions]{
\centering %图片居中
\includegraphics[width=0.3\textwidth]{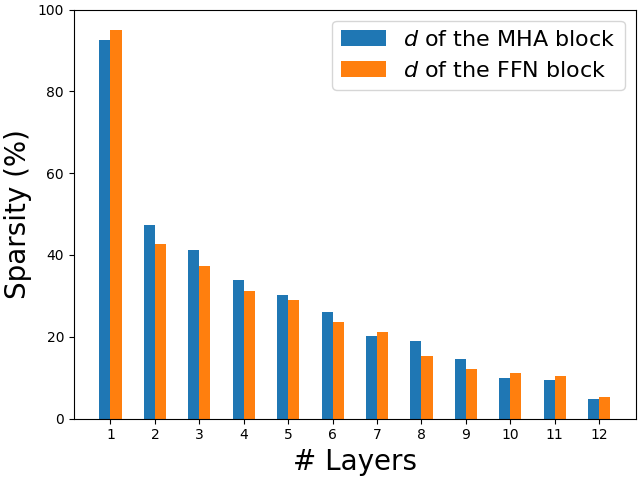} %插入图片，[]中设置图片大小，{}中是图片文件名
\label{Fig.Exp.1} %用于文内引用的标签  
}
\subfigure[Intermediate dimensions]{
\centering %图片居中
\includegraphics[width=0.3\textwidth]{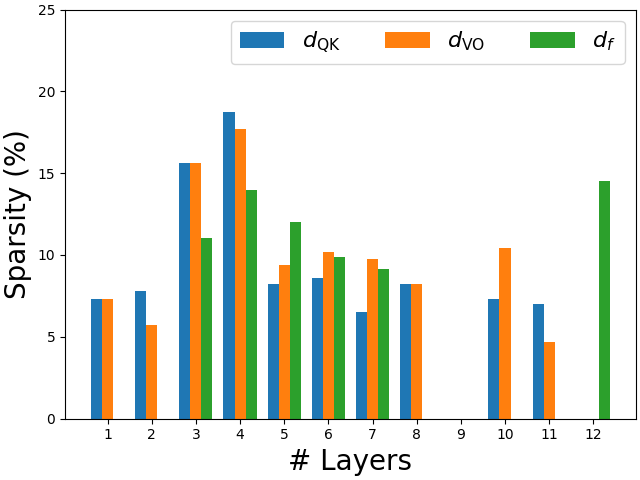} %插入图片，[]中设置图片大小，{}中是图片文件名
\label{Fig.Exp.2} %用于文内引用的标签  
}
\subfigure[Attention head size]{
\centering %图片居中
\includegraphics[width=0.3\textwidth]{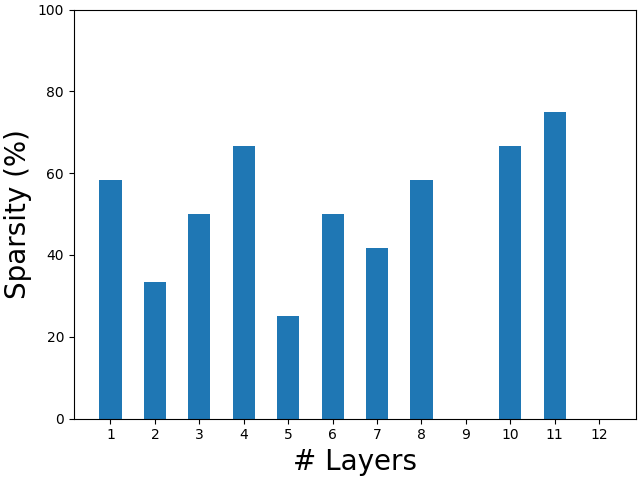} %插入图片，[]中设置图片大小，{}中是图片文件名
\label{Fig.Exp.3} %用于文内引用的标签  
}

\caption{\label{Fig.Exp} Structural information of the pruned model on the MRPC dataset, where sparsity denotes the ratio of the remaining dimension or size to the original dimension or size. (a) Output dimensions of each MHA and FFN block. (b) Intermediate dimensions of each MHA and FFN block. (c) The number of attention heads in each MHA block.} %最终文档中希望显示的图片标题
\end{figure*}
% -----------------------------------------------------------------------------------------

% -----------------------------------------------------------------------------------------
\begin{table*}[t]
\newcolumntype{?}{!{\vrule width 1pt}}
\newcolumntype{C}{>{\centering\arraybackslash}p{2em}}
\centering
\small

\renewcommand\arraystretch{1.2}

\resizebox{\linewidth}{!}{
\begin{tabular}{l|c|cccccccc|c}
\toprule
                  & Params.  & SST-2         & QNLI          & MNLI          & QQP           & RTE           & STS-B         & MRPC          & SQuAD  & Avg.       \\ \midrule
$\rm BERT_{base}$ & 85M      & 93.1          & 91.5          & 84.8          & 91.2          & 70.4          & 89.1          & 85.6          & 88.4   & 86.76      \\ \midrule
$\rm TinyBERT_4$  & 4.7M     & 89.7          & 86.7          & 78.8          & 90.0          & 63.2          & 85.0          & 81.4          & 82.1  &  82.11      \\
$\rm WID$         & 5.0M     & 88.8          & 85.4          & 78.4          & 89.5          & 60.3          & 84.5          & 81.9          & 81.2  &  81.25      \\
$\rm CoFi$        & \textasciitilde5.0M     & 90.6          & 86.1          & 80.6          & \textbf{90.1} & 64.7           & 83.1          & 82.6          & 82.6  &  82.55      \\ \midrule
$\rm SP^{3}$        & \textasciitilde5.0M     & \textbf{91.4} & \textbf{87.6} & \textbf{81.6} & \textbf{90.1} & \textbf{66.4} & \textbf{86.1} & \textbf{82.8} & \textbf{83.2} & \textbf{83.65}\\
Speedup          &         & $\times 2.65$ & $\times 2.79$ &  $\times 2.55$ &  $\times 2.55$ & $\times 4.98$ & $\times 5.58$  &  $\times 4.66$ & $\times 2.83$ &       \\
\bottomrule
\end{tabular}
}
\caption{Comparison between our \smodel and both the distillation methods and pruning methods. Note that, following CoFi, we do not count the number of parameters in the embedding layer.}
\label{Tab.baselines}
\end{table*}
% -----------------------------------------------------------------------------------------

% -----------------------------------------------------------------------------------------
\begin{table*}[t]
\newcolumntype{?}{!{\vrule width 1pt}}
\newcolumntype{C}{>{\centering\arraybackslash}p{2em}}
\centering
\small

\renewcommand\arraystretch{1.2}

\resizebox{0.9\linewidth}{!}{
\begin{tabular}{l|c|cccccccc|c}
\toprule
                  & Params.  & SST-2         & QNLI          & MNLI          & QQP           & RTE           & STS-B         & MRPC          & SQuAD  & Avg.       \\ \midrule
$\rm SP^3$        & 25M      & \textbf{91.1} & \textbf{89.6} & \textbf{80.8} & \textbf{90.9} & \textbf{66.8} & \textbf{86.8} & \textbf{85.0} & \textbf{84.8}     & \textbf{84.48}      \\
$\rm CoFi$    & 25M      & 86.9          & 82.2          & 78.7          & 90.0          & 56.7          & 82.3          & 71.8          & 79.1      & 78.46      \\
\bottomrule
\end{tabular}
}
\caption{Comparison of \smodel and CoFi in compressing the hidden dimension of $\rm BERT_{base}$.}
\label{Tab.compress_hidden_only}
\end{table*}
% -----------------------------------------------------------------------------------------

% -----------------------------------------------------------------------------------------
\begin{table}[t]
\newcolumntype{?}{!{\vrule width 1pt}}
\newcolumntype{C}{>{\centering\arraybackslash}p{2em}}
\centering
\small

\resizebox{\linewidth}{!}{
\begin{tabular}{lcccc}
\toprule
           & SST-2 & RTE  & STS-B & MRPC \\ \midrule
\model-10M     & \textbf{91.3}  & \textbf{66.8} & \textbf{87.8}  & \textbf{85.5} \\
\quad w/o PCA-Projection    & 91.2  & 52.7 & 85.0  & 80.6 \\ \midrule
\end{tabular}
}
\caption{Effect of PCA Projection. \model-10M refers to $\rm BERT_{base}$ compressed to 10M by omitting masks \( z_{\rm head} \), \( z_{\rm MHA} \), and \( z_{\rm FFN} \).}
\label{Tab.ablation_pca_projection}
\end{table}
% -----------------------------------------------------------------------------------------

% \model-5M     & \textbf{91.4}  & \textbf{66.4} & 86.1  & 82.8 \\
% \quad w/o head     & 90.9 & 66.1  & \textbf{86.6}  & 82.1     \\
% \quad w/o layer    & 91.2 & 62.8  & 86.1  & 83.5     \\ 
% \quad w/o head \& layer & 91.1 & 63.9 & 86.5  & \textbf{83.8}     \\ \bottomrule

% -----------------------------------------------------------------------------------------
\begin{table}[t]
\newcolumntype{?}{!{\vrule width 1pt}}
\newcolumntype{C}{>{\centering\arraybackslash}p{2em}}
\centering
\small

\resizebox{\linewidth}{!}{
\begin{tabular}{lcccc}
\toprule
Number of tokens    & SST-2          &  RTE            &  STS-B  & MRPC  \\ \midrule
\multicolumn{1}{c|}{2,048}  & 90.0           &  58.8           &  85.9   & 82.8  \\
\multicolumn{1}{c|}{4,096}  & \textbf{91.4}  &  \textbf{64.2}  &  86.1   & 82.8  \\ 
\multicolumn{1}{c|}{8,192}  & 90.8  &  63.5  &  \textbf{86.4}   & \textbf{83.1}  \\ \bottomrule
\end{tabular}
}
\caption{Effect of sampled tokens on PCA projection.}  % of PCA Projection
\label{Tab.ablation_num_tokens}
\end{table}
% -----------------------------------------------------------------------------------------

As shown in Table \ref{Tab.baselines}, we compress the $\rm BERT_{base}$ model and compare the performance of \smodel with other methods under similar sparsity. 
First, compared to the original model, \smodel retains 96\% of the model performance while removing 94\% of the model parameters. Second, compared to other baselines, \smodel has extra degrees of freedom. Compared to the pruning method CoFi, \smodel can additionally compress the hidden dimensions and intermediate dimensions of the MHA block. Unlike distillation methods, \smodel can use different hidden dimensions at different layers with additional head-level and layer-level pruning. Overall, \smodel obtains the best performance on all datasets, indicating that utilizing these extra degrees of freedom can benefit the performance during compression. The comparisons with additional baselines on $\rm BERT_{base}$ and experiments on TinyBERT with augmentation and OPT-125m are presented in Appendix \ref{Appendix.More_Results}. 

We delve into the impact of specifically compressing the hidden size \( d \) in our \smodel. This entails retaining only the projection matrices \( P_{\rm in}^{\cdot}, P_{\rm out}^{\cdot} \) and masks \( z_{\rm int}^{\cdot}, z_{\rm out}^{\cdot} \) in \smodel (refer to Eq.~\ref{Eq.hidden_dimension_pca_projection}), while maintaining the masks \( z_{\rm int}, z_{\rm out} \) in the CoFi approach (see Eq.~\ref{Eq.masked_base_pruning}). As indicated in Table \ref{Tab.compress_hidden_only}, \smodel demonstrates significant superiority over CoFi, which relies solely on masks for compressing the hidden dimensions. These experimental findings underscore the efficacy of PCA Projection in the compression of hidden dimensions, affirming its effectiveness in enhancing model performance.

We also apply our method to Large Language Models (LLMs), with experimental settings and results detailed in Appendix \ref{Appendix.sp3_for_llm}. Table~\ref{Tab.tinyllama} illustrates the effective compression achieved by our method on TinyLlaMa. Additionally, thanks to the pre-RMSNorm architecture of TinyLlaMa, we propose Group PCA Projection. This approach unifies the computation of main components of features across several consecutive blocks, effectively reducing parameters introduced at the residuals and improving performance compared to distinct main components in each block, as demonstrated in Table~\ref{Tab.tinyllama-abl-gs}.

\subsection{Ablation Study on PCA Projection}

% obtained through SVD decomposition

% head-level and layer-level

PCA Projection employs principal components as the projection matrix. To assess the importance of these principal components, we conduct an experiment where we initialize the projection matrices as identity matrices. To prevent other masks from interfering with the role of PCA Projection, we use only dimension-level masks in our experiments. The results, as depicted in Table \ref{Tab.ablation_pca_projection}, indicate that excluding PCA Projection results in a significant performance decline on the RTE and MRPC datasets, with a slight decrease observed on the SST-2 and STS-B datasets. This is because, for larger datasets, the model can directly optimize suitable compactor parameters from the data. The results underscore the significant role of PCA Projection in enhancing the performance of the compressed model.

Furthermore, PCA Projection relies on calibration data derived from the training dataset. We examine how the number of tokens in this data affects model performance. As shown in Table \ref{Tab.ablation_num_tokens}, the accuracy of the model varies with different sample token sizes. Considering that a higher number of samples adds complexity to the SVD computation, choosing 4,096 tokens strikes an optimal balance between accuracy and computational efficiency.
We also conduct experiments using different random seeds to sample calibration data and show its robustness. Appendix \ref{Appendix.more_ablation} presents the details as well as the studies about different masks.

% Refer to Appendix \ref{Appendix.more_ablation} for more ablation experiments on head- and layer-level masks.
% The ablation studies on 

\subsection{Structures of Pruned Model}

We study the pruned structures produced by \model. Take the MRPC dataset as an example. Figure \ref{Fig.Exp} shows the structural information of the pruned model. More results are shown in Appendix \ref{Appendix.Structures}. 

From Fig. \ref{Fig.Exp} (b) and (c), as well as figures for other datasets, it's evident that the model structure varies across different datasets. However, a consistent observation across these structures is that layers nearer the output are more compressed than those closer to the inputs. Additionally, the intermediate dimensions of the FFN block are notably more compressed across all datasets compared to the intermediate dimensions of the MHA block. This distinction is highlighted when comparing the green bars to the blue and red bars in Fig. \ref{Fig.Exp} (b). The compression patterns we observe are consistent with those reported in prior pruning studies~\cite{cofi.xia2022}. In addition, Fig. \ref{Fig.Exp} (a) and similar figures for other datasets indicate a trend where the model's hidden dimension reduces as the number of layers increases. This trend suggests that the model incrementally compresses features into more compact dimensions during its inference process.

%% file: contents/relatedwork.tex
\section{Related Work}

\vpara{Distillation.} Knowledge distillation \cite{distilling.hinton2015, Dynabert.hou2020, TinyBERT.jiao2020, Distillbert.VictorSanh} is a model compression approach that transfers knowledge from a larger teacher model to a smaller student model. In contrast to the distillation methods, our approach uses less computational cost while achieving the same performance.

% Pre-training of the student model is important for most distillation methods, but this results in high computational costs~\cite{}. 
% from scratch on unlabeled corpora
% The advantage of distillation is that the structure of the student model can be specified arbitrarily.
% Some distillation methods, such as DistillBERT \cite{Distillbert.VictorSanh}, initialize the student model through the teacher model to avoid the pre-training phase, but these methods limit the possible structures of the student model.
% WID \cite{WID.wu2023} inherits the parameters of the teacher model and tries to directly compress the teacher model into the student model through the re-parameterization method.

\vpara{Pruning.} Existing pruning methods can be broadly divided into two categories: unstructured and structured. Unstructured pruning \cite{State.gale2019, lottery.jonathan2018, Optimal.kurtic2022, l0pruning.louizos2018, Movement.sanh2020} aims to remove unimportant scalar values from the model's parameters. In contrast, structured pruning \cite{Slip.lin2020, BMP.lagunas2021, Structured.wang2020, cofi.xia2022} are proposed to remove weight blocks in PLMs, including the entire layer \cite{Reducing.fan2019, DropLayer.sajjad2023, Poor.sajjad2020}, attention heads \cite{HeadPruning.michel2019, HeadPruning.voita2019}, and filters \cite{FFNPruning.mccarley2019, FFNPruning.prasanna2020}. Unlike the previous structured pruning, we focus on exploring how to compress the hidden dimensions of the model efficiently.

 % Although unstructured pruning algorithms can remove many redundant parameters while ensuring accuracy, compressed models require specific sparse data structures and hardware support to take advantage of unstructured pruning. For this reason

% \vpara{Low-Rank Factorization.} Low-rank factorization methods compress the PLMs by decomposing the weight matrices \cite{tucker.liu2022, Tensorized.ma2019, TFWSVD.hua2022, COMCAT.xiao2023, tucker.yin2022, CPTensor.zhou2019} based on low-rank property. In our approach, we propose PCA Projection based on the low-rank property of the features.

\vpara{Low-Rank Factorization (LRF).} LRF methods compress the PLMs by decomposing the weight matrices \cite{TFWSVD.hua2022, COMCAT.xiao2023} based on low-rank property. In contrast, we propose PCA Projection based on the low-rank property of the features. For more discussions refer to Appendix \ref{Appendix.comparison_with_low_rank}.

% Other works \cite{Tensorized.ma2019, COMCAT.xiao2023} have considered the model structure of PLM while performing matrix decomposition, and these works are mainly used for the compression of MHA blocks.

\vpara{Re-parameterization (Re-p).} Re-p methods have been proposed to improve existing model compression methods~\cite{resrep.ding2021, weight-distill.lin-etal-2021, WID.wu2023}. Its core idea is to represent the compressed model based on the original model. Therefore, these methods can utilize the parameters of the model more efficiently, which leads to better performance. \smodel can also be regarded as a Re-p method, and the major difference between \smodel and the existing methods is that we better initialize the parameters based on the PCA theory.

% \vpara{Re-parameterization (Re-p).} Re-p methods have been proposed to improve existing model compression methods~\cite{resrep.ding2021, weight-distill.lin-etal-2021, WID.wu2023}. Its core idea is to represent the compressed model based on the original model. Therefore, this method can utilize the parameters of the original model more efficiently, which leads to better performance of the compressed model. \smodel can also be regarded as a Re-p method, and the major difference between \smodel and the existing methods is that we better initialize the parameters additionally inserted into the model by the Re-p method based on the PCA theory.

%% file: contents/conclusion.tex
\section{Conclusion}

This study introduces \model, an enhanced structured pruning approach for compressing PLMs. \smodel employs PCA Projection, facilitating easier optimization of the compression model, thereby boosting its performance. Concurrently, \smodel uses non-shared masks in the hidden dimension. This allows the model to selectively determine its hidden dimensions size at different layers in line with the desired sparsity. When applied to $\rm BERT_{base}$ and evaluated on the GLUE and SQuAD benchmarks, \smodel notably achieves a 94\% sparsity with only a minor 4\% reduction in accuracy.

%% file: appendix/MHA_pca_projection.tex
\section{MHA Intermediate Dimension PCA Projection}
\label{Appendix.MHA_PCA_Projection}

We can verify the effectiveness of the Eq. \ref{Eq.MHA_PCA_projection} in the following way, 

\begin{small}
\begin{align}
\begin{split}
    \Vert X^{\top} W_Q^{(i)\top} W_K^{(i)} X - X^{\top} W_Q^{(i)\top} U_{Q,:k}^{(i)\top} U_{K,:k}^{(i)} W_K^{(i)} X \Vert,
    \label{Eq.metric}
\end{split}
\end{align}
\end{small}

\noindent Eq. \ref{Eq.metric} measures the difference between the inner product computed using the first $k$ principal components of $U_Q^{(i)}, U_K^{(i)}$ and the original result.

Next, we prove that $U_Q^{(i)}$ and $ U_K^{(i)}$ satisfy the following equation,

\begin{small}
\begin{align}
    {\rm M}_k = U_{Q,:k}^{(i)\top} U_{K,:k}^{(i)}
\end{align}
\end{small}

\noindent where,

\begin{small}
\begin{align}
\begin{split}
    {\rm M}_k = \ & \underset{{\rm M}}{\arg\min} \Vert X^{\top} W_Q^{(i)\top} W_K^{(i)} X - X^{\top} W_Q^{(i)\top} {\rm M} W_K^{(i)} X \Vert, \\
                & {\rm s.t.} \ rank({\rm M}) = k
\end{split}
\end{align}    
\end{small}

\noindent Suppose 

\begin{small}
\begin{align}
\begin{split}
& X_Q = W_Q^{(i)}X, \\
& X_K = W_K^{(i)}X, \\
& U_1, \Sigma_1, V_1^{\top} = {\rm SVD}(X_Q), \\
& U_2, \Sigma_2, V_2^{\top} = {\rm SVD}(X_K).
\end{split}
\end{align}
\end{small}

\noindent Then, we have

\begin{small}
\begin{align}
\begin{split}
    {\rm M}_k & =  \underset{{\rm M}}{\arg\min} \Vert X_Q^{\top} X_K - X_Q^{\top} {\rm M} X_K \Vert \\
              & =  \underset{{\rm M}}{\arg\min} \Vert V_1^{\top} X_Q^{\top} X_K V_2 -  V_1^{\top} X_Q^{\top} {\rm M} X_K V_2 \Vert \\
              & =  \underset{{\rm M}}{\arg\min} \Vert \Sigma_1^{\top} U_1^{\top} U_2 \Sigma_2 - \Sigma_1^{\top} U_1^{\top} {\rm M} U_2 \Sigma_2 \Vert
\end{split}
\end{align}    
\end{small}

\noindent Define

\begin{small}
\begin{align}
\begin{split}
    & Z  = \Sigma_1^{\top} U_1^{\top} U_2 \Sigma_2, \\
    & U_Z, \Sigma_Z, V_Z^{\top} = {\rm SVD}(Z),
\end{split}
\end{align}
\end{small}

\noindent then, based on the PCA theory, we know that {\small $Z_k = U_{Z,:k} \Sigma_{Z,:k} V_{Z,:k}^{\top}$} is the solution of 

\begin{small}
\begin{align}
    \underset{Z^*}{\arg\min} \Vert Z - Z^* \Vert, \quad {\rm s.t.} \ {\rm rank}(Z^*) = k.    
\end{align}
\end{small}

\noindent Therefore, we have

\begin{small}
\begin{align}
\begin{split}
    & \Sigma_Q^{\top} U_1^{\top} {\rm M_k} U_2 \Sigma_2 = U_{Z,:k} \Sigma_{Z,:k} V_{Z,:k}^{\top} \\    
    & \Rightarrow {\rm M}_k = U_1 \Sigma_1^{-1\top} U_{Z,:k} \Sigma_{Z,:k} V_{Z,:k}^{\top} \Sigma_2^{-1} U_2^{\top}.
\end{split}
\end{align}
\end{small}

\noindent Refer to Eq. \ref{Eq.MHA_PCA_projection}, 

\begin{small}
\begin{align}
\begin{split}
U_Q^{(i)} & = \Sigma_{Z}^{(i)\frac{1}{2}} U_{Z}^{(i)\top} \Sigma_1^{(i)-1} U_1^{(i)\top}, \\
U_K^{(i)} & = \Sigma_{Z}^{(i)\frac{1}{2}} V_{Z}^{(i)} \Sigma_2^{(i)-1} U_2^{(i)\top}, 
\end{split}
\end{align}    
\end{small}

\noindent we also have

\begin{small}
\begin{align}
\begin{split}
U_{Q,:k}^{(i)} & = \Sigma_{Z,:k}^{(i)\frac{1}{2}} U_{Z,:k}^{(i)\top} \Sigma_1^{(i)-1} U_1^{(i)\top}, \\
U_{K,:k}^{(i)} & = \Sigma_{Z,:k}^{(i)\frac{1}{2}} V_{Z,:k}^{(i)} \Sigma_2^{(i)-1} U_2^{(i)\top}.
\end{split}
\end{align}    
\end{small}

\noindent Therefore,

\begin{small}
\begin{align}
    {\rm M}_k = U_{Q,:k}^{(i)\top} U_{K,:k}^{(i)}
\end{align}
\end{small}

\noindent

%% file: appendix/pseudo_code.tex
\section{Pseudo Code of PCA Projection}
\label{Appendix.code}

The pseudo-code of PCA Projection is shown in Algorithm \ref{alg:pca_projection}.
\begin{algorithm*}[t]
\small
\caption{PCA Projection} \label{alg:pca_projection}
\KwIn{$T$: sampled tokens size, $\mathbb{D}$: calibration data, $L^{(1 \sim N)}$: model }
\KwOut{Projection Matrix $P_{\rm in}, P_{\rm out}, P_Q, P_K, P_V, P_O$}

$X_{\rm Q}, X_{\rm K}, X_{\rm V}, X_{\rm M}, X_{\rm F} \leftarrow [], [], [], [], []$ \\

\For{$j \leftarrow 1$ \KwTo $|\mathbb{D}|$}{
    $x^{(0)} \leftarrow \mathbb{D}[j]$ \\
    \For{$i \leftarrow 1$ \KwTo $N$}{
        $[W_Q, W_K, W_V, W_O] \leftarrow L^{(i)}$ \\
        \For{$h \leftarrow 1$ \KwTo $d_h$}{
            $X_Q^{(h, i)} \leftarrow {\rm Concate}(X_Q^{(h, i)}, W_Q^{(h)} x^{(i - 1)})$ \\
            $X_K^{(h, i)} \leftarrow {\rm Concate}(X_K^{(h, i)}, W_Q^{(h)} x^{(i - 1)})$ \\
            $X_V^{(h, i)} \leftarrow {\rm Concate}(X_V^{(h, i)}, W_Q^{(h)} x^{(i - 1)})$ \\
        }
        $x_{\rm M} = {\rm MHA}^{(i)}(x^{(i-1)}) + x^{(i-1)}$ \\
        $x_{\rm M}^{\rm LN} = {\rm LN}(x_{\rm M})$ \\
        $x_{\rm F} = {\rm FFN}^{(i)}(x_{\rm M}) + x_{\rm M}$ \\
        $x_{\rm F}^{\rm LN} = {\rm LN}(x_{\rm F})$ \\     
        $x^{(i)} = x_{\rm F}^{\rm LN}$ \\
        $X_{\rm M}^{(i)} \leftarrow {\rm Concate}(X_{\rm M}^{(i)}, x_{\rm M})$ \\
        $X_{\rm F}^{(i)} \leftarrow {\rm Concate}(X_{\rm F}^{(i)}, x_{\rm F})$ \\
    }
}
\For{$i \leftarrow 1$ \KwTo $N$}{
    $[W_Q, W_K, W_V, W_O] \leftarrow L^{(i)}$ \\
    \For{$h \leftarrow 1$ \KwTo $d_h$}{
        $X_V^{(h, i)} \leftarrow {\rm Sample}(X_V^{(h, i)}, T)$ \\
        $X_Q^{(h, i)} \leftarrow {\rm Sample}(X_Q^{(h, i)}, T)$ \\
        $X_K^{(h, i)} \leftarrow {\rm Sample}(X_K^{(h, i)}, T)$ \\

        $U_V, \Sigma_V, V^\top_V \leftarrow {\rm SVD}(X_V^{(h, i)})$ \\
        $U_1, \Sigma_1, V^\top_1 \leftarrow {\rm SVD}(X_Q^{(h, i)})$ \\
        $U_2, \Sigma_2, V^\top_2 \leftarrow {\rm SVD}(X_K^{(h, i)})$ \\
        $Z = \Sigma_1^{\top} U_1^{\top} U_2 \Sigma_2$ \\
        $U_Z, \Sigma_Z, V_Z^{\top} = {\rm SVD}(Z)$ \\
        $ U_Q = \Sigma_{Z}^{\frac{1}{2}} U_{Z}^{\top} \Sigma_1^{-1} U_1^{\top} $ \\
        $ U_K = \Sigma_{Z}^{\frac{1}{2}} V_{Z} \Sigma_2^{-1} U_2^{\top} $ \\

        $P_V^{(h, i)}, P_O^{(h, i)} \leftarrow U_V^{\top}, U_V$ \\
        $P_Q^{(h, i)}, P_K^{(h, i)} \leftarrow U_Q, U_K$ \\

    }
    $[\gamma^M, \gamma^F] \leftarrow L^{(i)}$ \\
    $R \leftarrow I - \frac{1}{d}11^\top$ \\
    $X_{\rm M}^{(i)} \leftarrow {\rm Sample}(X_{\rm M}^{(i)}, T)$ \\
    $X_{\rm F}^{(i)} \leftarrow {\rm Sample}(X_{\rm F}^{(i)}, T)$ \\
    $U_{\rm M}, \Sigma_{\rm M}, V^\top_{\rm M} \leftarrow {\rm SVD}(X_{\rm M}^{(i)})$ \\
    $U_{\rm F}, \Sigma_{\rm F}, V^\top_{\rm F} \leftarrow {\rm SVD}(X_{\rm F}^{(i)})$ \\
    $P_{\rm in}^{({\rm M}, i)}, P_{\rm out}^{({\rm M}, i)} \leftarrow U_{\rm M}^\top R, \, {\rm diag}(\gamma^{\rm M}) U_{\rm M}$ \\
    $P_{\rm in}^{({\rm F}, i)}, P_{\rm out}^{({\rm F}, i)} \leftarrow U_{\rm F}^\top R, \, {\rm diag}(\gamma^{\rm F}) U_{\rm F}$ \\
}

\end{algorithm*}

%% file: appendix/sparsity.tex
\section{Sparsity}
\label{Appendix.Sparsity}

The expected sparsity $\hat{s}$ is computed as follows:

\begin{small}
\begin{align}
\begin{split}
    & \hat{s} = \frac{1}{M} ( \\
    & \sum_{i}^{L} \sum_{j}^{H} \sum_{k}^{d} \sum_{l}^{d_h} z_{\rm MHA}^{(i)} \cdot z_{\rm head}^{(i,j)} \cdot z_{\rm out, FFN, i-1}^{(i,k)} \cdot z_{Q}^{(i,l)} + \\
    & \sum_{i}^{L} \sum_{j}^{H} \sum_{k}^{d} \sum_{l}^{d_h} z_{\rm MHA}^{(i)} \cdot z_{\rm head}^{(i,j)} \cdot z_{\rm out, FFN, i-1}^{(i,k)} \cdot z_{K}^{(i,l)} + \\
    & \sum_{i}^{L} \sum_{j}^{H} \sum_{k}^{d} \sum_{l}^{d_h} z_{\rm MHA}^{(i)} \cdot z_{\rm head}^{(i,j)} \cdot z_{\rm out, FFN, i-1}^{(i,k)} \cdot z_{V}^{(i,l)} + \\
    & \sum_{i}^{L} \sum_{j}^{H} \sum_{k}^{d} \sum_{l}^{d_h} z_{\rm MHA}^{(i)} \cdot z_{\rm head}^{(i,j)} \cdot z_{\rm in, MHA, i}^{(i,k)} \cdot z_{O}^{(i,l)} + \\
    & \sum_{i}^{L} \sum_{k}^{d} \sum_{l}^{d_f} z_{\rm FFN}^{(i)} \cdot z_{\rm out, MHA, i}^{(i,k)} \cdot z_{f}^{(i,l)} + \\
    & \sum_{i}^{L} \sum_{k}^{d} \sum_{l}^{d_f} z_{\rm FFN}^{(i)} \cdot z_{\rm in, FFN, i}^{(i,k)} \cdot z_{f}^{(i,l)} + \\
    & \sum_{i}^{L} \sum_{k}^{d} z_{\rm out, FFN, i-1}^{(i,k)} \cdot z_{\rm in, MHA, i}^{(i,k)} + \\
    & \sum_{i}^{L} \sum_{k}^{d} z_{\rm out, MHA, i}^{(i,k)} \cdot z_{\rm in, FFN, i}^{(i,k)}),
\end{split}
\end{align}
\end{small}

\noindent where $M$ denotes the total number of parameters of the model. 

%% file: appendix/experiment.tex
\section{Experiment Setting Details}
\label{Appendix.Experiment}

\subsection{Datasets}
\label{appendix.dataset}

GLUE \cite{GLUE.wang2018} benchmark consists of various tasks related to sentence similarity calculation, classification, textual entailment, and natural language inference. It includes 10 tasks, namely AX, COLA, QQP, MNLI, MRPC, QNLI, QQP, RTE, SST-2, STS-B, and WNLI. The number of training examples for each task is as follows: 1.1k, 10.7k, 432k, 5.8k, 105k, 364k, 3k, 70k, 67k, and 852, respectively. SQuAD 1.1 \cite{SQuAD.2016} dataset involves question answering tasks, containing 88K training examples.

\subsection{Experiment Setup}
\label{appendix.setup}

Our \smodel is developed using PyTorch \cite{pytorch} and executed on a server equipped with four NVIDIA 3090 GPU cards for all experiments. We source the BERT model from the HuggingFace \cite{huggingface} Transformers library. To prepare for compression, we first fine-tune BERT to create a task-specific model. This fine-tuning occurs on the training datasets from the GLUE benchmark and SQuAD, spanning 3 epochs, with a batch size of 32 and learning rates of 1e-5 and 2e-5, respectively. We maintain the default hyperparameters provided by HuggingFace.

The pruning of BERT follows the same configuration as the model fine-tuning process. We first fine-tune \smodel with the task-specific objective for 2 epochs, then dimension-level pruning begins at the 3rd epoch, followed by the head-level and layer-level pruning at the 8th epoch.

%% file: appendix/more_result.tex
\section{More Experimental Results}
\label{Appendix.More_Results}

\begin{table*}[t]
\newcolumntype{?}{!{\vrule width 1pt}}
\newcolumntype{C}{>{\centering\arraybackslash}p{2em}}
\centering
\small

\renewcommand\arraystretch{1.2}

\resizebox{\linewidth}{!}{
\begin{tabular}{l|c|cccccccc}
\toprule
                  & Params.  & SST-2         & QNLI          & MNLI          & QQP           & RTE           & STS-B         & MRPC          & SQuAD       \\ \midrule
$\rm BERT_{base}$ & 110M      & 93.1          & 91.5          & 84.8          & 91.2          & 70.4          & 89.1          & 85.6          & 88.4         \\ \midrule
$\rm SP^{3}$                              & 25M  & 91.4 & 87.6 & 81.6 & 90.1 & 66.4 & 86.1 & 82.8 & 83.2 \\ \midrule
DisitilBERT \cite{Distillbert.VictorSanh} & 13.6M & 86.4 & 84.5 & 71.3 & 88.0 & 56.3 & -    & 72.5 & 56.2 \\
LPAF \cite{LPAF.ren2023}                  & 13.6M & 89.7 & 88.6 & 81.7 & 90.1 & 67.9 & -    & 86.0 & 75.1 \\
LRC-BERT \cite{lrc-bert.fu2021}           & 14.5M & 92.9 & 88.7 & 82.7 & 72.2 & 63.1 & 81.2 & 87.9 & -    \\
AD-KD \cite{AD-KD.wu2023}                 & 66M   & 91.8 & 90.0 & 82.6 & 88.9 & 65.8 & 83.4 & 87.1 & -    \\
DynaBERT \cite{Dynabert.hou2020}          & 10.6M & 92.0 & 88.5 & 82.3 & 90.4 & 63.2 & 87.0 & 81.4 & 76.6 \\
BERT-EMD \cite{bert-emd.li2020}           & 66M   & 93.3 & 90.7 &  84.7 & 72.0 & 71.7 & 86.8 & 89.8 & -   \\
KroneckerBERT \cite{Kroneckerbert.tahaei2022} & 5.2M & 88.4 & 86.1 & 80.1 & 70.5 & 64.7 & 81.3 & 87.1 & - \\
BERT-of-Theseus \cite{Bert-of-theseus.xu2020} & 66M   & 91.5 & 89.5 & 82.3 & 89.6 & 68.2 & 88.7 & 89.0 & - \\ \midrule
Movement \cite{Movement.sanh2020}         & 3M   &  -   &  -   &  79.5   &  89.1   &  -   &  -   &  -   &  82.3  \\
FLOP \cite{Structured.wang2020}           & 80M   & 92.1 & 89.1 & -    & -    & -    & 88.2 & 88.6 & -    \\
ROSITA \cite{rosita.liu2021}              & 14.5M & 87.6 & 83.8 & 77.7 & 88.3 & -    & -    & -    & -    \\
PLATON \cite{platon.zhang2022}            & 8.5M  & 90.5 & 88.9 & 82.2 & 90.2 & 65.3 & 87.1 & 84.3 & 79.0 \\
CAP-f \cite{CAP-f.xu2022}                 & 8.5M  & 89.7 & -    & 81.2 & 90.2 & -    & -    & -    & 70.2 \\
Fast \cite{Fast.kwon2022}                 & 66M & 92.5 & 90.1 & 82.5 & 90.4 & -    & 88.0 & 85.3 & 75.3 \\
KCM \cite{KCM.nova2023}                   & 50M   & 91.1 & 87.8 & 77.2 & 89.2 & -    & 85.7 & 84.2 & 70.3 \\
\bottomrule
\end{tabular}
}
\caption{Comparison between our \smodel and other distillation and pruning methods on $\rm BERT_{base}$.}
\label{Tab.more_baselines}
\end{table*}

\begin{table*}[t]
\newcolumntype{?}{!{\vrule width 1pt}}
\newcolumntype{C}{>{\centering\arraybackslash}p{2em}}
\centering
\small

\renewcommand\arraystretch{1.2}

% SST-2	QNLI	MRPC	RTE

\resizebox{0.5\linewidth}{!}{
\begin{tabular}{l|cccc}
\toprule
                & SST-2  & QNLI  & MRPC       & RTE \\ \midrule
TinyBERT w/  DA & \textbf{91.6}   & \textbf{87.6}  & \textbf{83.6}       & 62.5 \\ 
TinyBERT w/o DA & 89.7   & 86.7  & 81.4       & 63.2 \\
$\rm SP^3$      & 91.4   & \textbf{87.6}  & 82.8       & \textbf{66.4} \\
\bottomrule
\end{tabular}
}
\caption{Comparison between our \smodel on $\rm BERT_{base}$ and TinyBERT, where DA means data augmentation. }
\label{Tab.compare_with_tinybert}
\end{table*}

\begin{table*}[t]
\newcolumntype{?}{!{\vrule width 1pt}}
\newcolumntype{C}{>{\centering\arraybackslash}p{2em}}
\centering
\small

\renewcommand\arraystretch{1.2}

\resizebox{0.6\linewidth}{!}{
\begin{tabular}{l|c|cccccc}
\toprule
                 & Params.  & SST-2         & QNLI          & QQP           & RTE           & STS-B         & MRPC       \\ \midrule
$\rm OPT_{125m}$ & 85M      & 92.9          & 90.8          & 90.1          & 66.4          & 87.0          & 82.4       \\ \midrule
$\rm CoFi^*$     & 7M       & 86.8          & 84.4          & \textbf{88.5} & 55.8          & 70.6          & 69.2       \\ 
$\rm SP^{3}$     & 7M       & \textbf{89.9} & \textbf{86.7} & 87.8          & \textbf{60.3} & \textbf{75.6} & \textbf{77.2}       \\ 

\bottomrule
\end{tabular}
}
\caption{Pruning performance of \smodel on $\rm OPT_{125m}$, where $\rm CoFi^*$ means that $\rm CoFi$ on $\rm OPT_{125m}$ is implemented by ourselves. }
\label{Tab.opt125m}
\end{table*}

\begin{table*}[t]
\newcolumntype{?}{!{\vrule width 1pt}}
\newcolumntype{C}{>{\centering\arraybackslash}p{2em}}
\centering
\small

\renewcommand\arraystretch{1.2}

\resizebox{0.9\linewidth}{!}{
\begin{tabular}{l|c|ccccccc}
\toprule
                 & Params.        & ARC-e  & ARC-c       & BoolQ    & HellaSwag  & OBQA    & PIQA  & WinoGrande      \\ \midrule
$\rm TinyLlaMa$  & 1.1B           & 57.5   & 27.4        & 59.2     & 44.0       & 23.8    & 70.8  & 56.0            \\ \midrule
$\rm SP^3$       & 0.85B          & 51.6   & 25.9        & 50.5     & 37.5       & 20.4    & 67.3  & 54.3            \\ 
                 & 0.65B          & 44.9   & 22.3        & 61.0     & 33.0       & 17.2    & 61.3  & 53.7            \\ 
\bottomrule
\end{tabular}
}
\caption{Pruning performance of \smodel on TinyLlaMa. }
\label{Tab.tinyllama}
\end{table*}

\begin{table*}[t]
\newcolumntype{?}{!{\vrule width 1pt}}
\newcolumntype{C}{>{\centering\arraybackslash}p{2em}}
\centering
\small

\renewcommand\arraystretch{1.2}

\resizebox{0.9\linewidth}{!}{
\begin{tabular}{l|c|ccccccc}
\toprule
                        & Params.  & ARC-e  & ARC-c       & BoolQ    & HellaSwag  & OBQA    & PIQA  & WinoGrande      \\ \midrule
$\rm SP^3$-gs1          & 0.85B + 0.10B   & 38.5   & \textbf{23.6}        & 51.0     & 30.0       & 14.8    & 57.9  & 50.8            \\ 
$\rm SP^3$-gs2          & 0.85B + 0.05B  & 42.0   & 21.7        & 52.1     & 31.7       & 17.0    & 60.2  & \textbf{55.6}            \\ 
$\rm SP^3$-gs4          & 0.85B + 0.03B  & 44.2   & 23.4        & 58.3     & 33.0       & \textbf{18.4}    & 61.5  & 52.6            \\ 
$\rm SP^3$-gs8          & 0.85B + 0.01B  & \textbf{46.0}   & \textbf{23.6}        & \textbf{59.0}     & \textbf{33.6}       & 17.6    & \textbf{62.0}  & 54.0            \\ 
\bottomrule
\end{tabular}
}
\caption{Pruning performance  of $\rm SP^3$ on TinyLlaMa under different group sizes (gs).}
\label{Tab.tinyllama-abl-gs}
\end{table*}

\begin{table*}[t]
\newcolumntype{?}{!{\vrule width 1pt}}
\newcolumntype{C}{>{\centering\arraybackslash}p{2em}}
\centering
\small

\renewcommand\arraystretch{1.2}

\resizebox{0.9\linewidth}{!}{
\begin{tabular}{l|c|ccccccc}
\toprule
                        & Params.  & ARC-e  & ARC-c       & BoolQ    & HellaSwag  & OBQA    & PIQA  & WinoGrande      \\ \midrule
LLM-Pruner              & 0.85B    & 46.2   & 23.1        & 45.2     & 35.4       & 16.8    & 65.6  & \textbf{53.7}            \\ 
LLM-Pruner + $\rm SP^3$ & 0.85B    & \textbf{47.3} & \textbf{23.3} & \textbf{48.4} & \textbf{35.9} & \textbf{17.0} & \textbf{68.0} & 52.0            \\ 
\bottomrule
\end{tabular}
}
\caption{Pruning performance  of combining $\rm SP^3$ and LLM-Pruner on TinyLlama. }
\label{Tab.tinyllama-llm-pruner}
\end{table*}

\subsection{Comparing \smodel with More Baselines}

The comparison results between \smodel and more baselines are shown in Table \ref{Tab.more_baselines}.

\subsection{Comparing \smodel with TinyBERT}

We skip the data augmentation step in TinyBERT, following the approach of CoFi \cite{cofi.xia2022}, as this data augmentation significantly enlarges the training dataset, leading to a substantial increase in training time. Here, we just present the results of TinyBERT with data augmentation (DA) for reference. The comparison results are shown in Table \ref{Tab.compare_with_tinybert}. Even with data augmentation, our proposed $\rm SP^3$ demonstrates comparable performance. 

\subsection{Experiment Results on $\rm OPT_{125m}$}

The results of the experiment on $\rm OPT_{125m}$ are shown in Table \ref{Tab.opt125m}. For $\rm OPT_{125m}$, we use the same experimental setup as for $\rm BERT_{base}$.

% \subsection{Experiment Results on TinyLlama}

% The results of the experiment on TinyLlama (1.1B) \cite{tinyllama} are shown in Table \ref{Tab.tinyllama}. we use the SFT dataset oasst\_top1\_2023-08-25 to prune the TinyLlama. We prune the model on the oasst\_top1\_2023-08-25 dataset for 2 epochs. Then, we tested the performance of the compressed model under a zero-shot setting.

%% file: appendix/SP3_for_LLM.tex
\section{$\rm SP^3$ for Large Language Model}
\label{Appendix.sp3_for_llm}

\begin{figure}[t]
    \centering
    \includegraphics[width=0.5\textwidth]{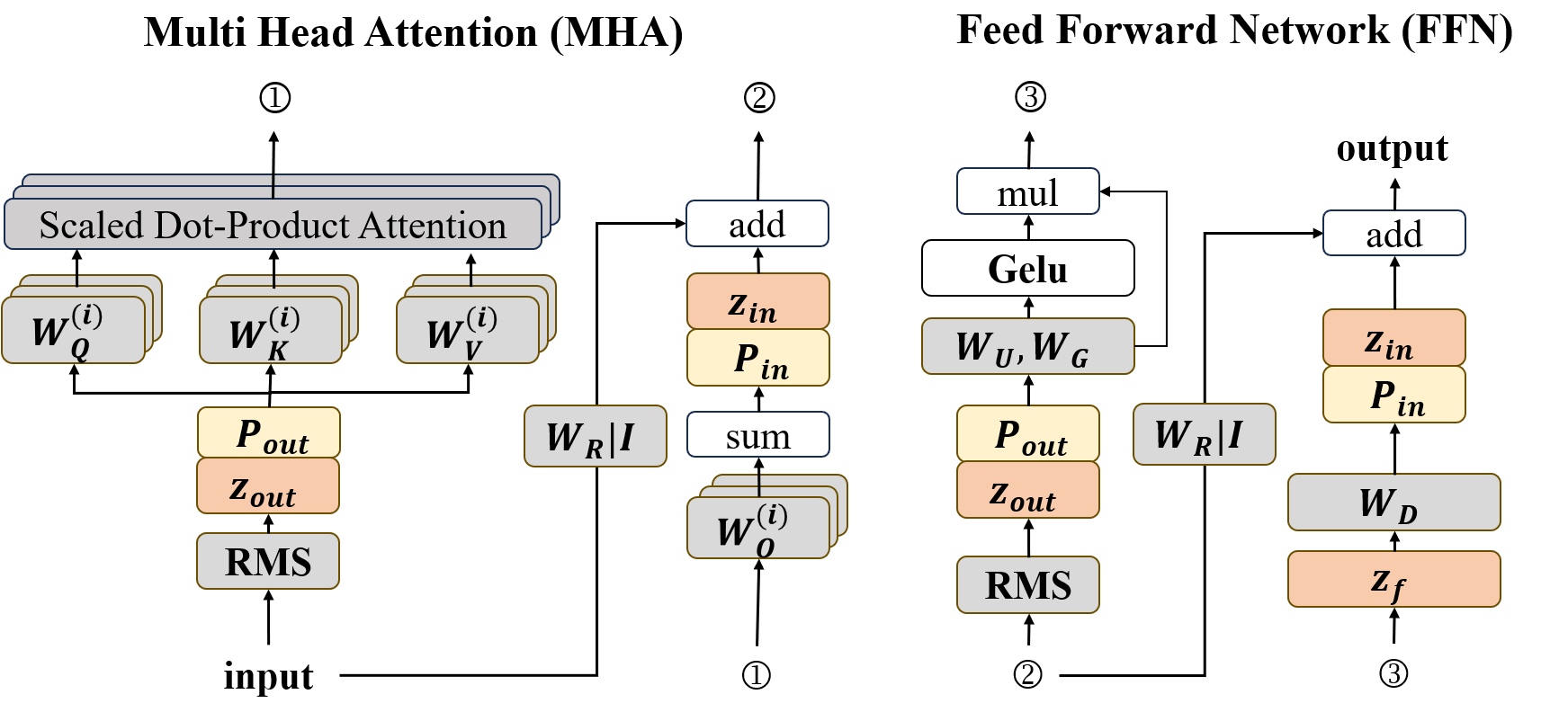}
    \caption{Illustration of the \smodel architecture for LLM, in which the gray rectangles represent the weight matrices, the yellow rectangles signify the projection matrices, and the red rectangles indicate the masks.}
    \label{Fig.model-structure-llama}
\end{figure}

$\rm SP^3$ can also be used to prune large language model (LLM), e.g., Llama \cite{llama.touvron2023}). To mitigate computational overhead of LLM, our approach restricts using SFT data exclusively for model pruning. Concurrently, to reduce pruning complexities, we only prune the hidden dimensions and the filters within the FFN block, while ignoring the pruning of attention heads and layers, because coarse-grained pruning requires more training data to recover the performance of the model. We also ignore the pruning of the intermediate dimensions of the MHA block, as it may conflict with the RoPE positional encoding \cite{rope.su2024} which is commonly used for LLM. The structure used to prune the LLM is shown in Fig. \ref{Fig.model-structure-llama}.

\subsection{Group PCA Projection}

As shown in section \ref{section:fusing}, to compress the hidden dimensions of the model, we need to add additional matrices to each residual. These additional added matrices increase the total number of parameters in the model by 15\%. Fortunately, we find that if we use the same principal components to initialize the compactors of successive blocks, we can avoid introducing additional matrices at the residuals. Taking Llama as an example, for the $i$-th block of the Llama model, we have

\begin{small}
\begin{align}
    x^{(i)} & = {\rm Block}^{(i)}({\rm RMSNorm}(x^{(i-1)})) + x^{(i-1)}.
\end{align}    
\end{small}

\noindent After computing the principal component matrix $U$, we have

\begin{small}
\begin{align}
    x^{(i)} & = {\rm Bloc}k^{(i)}({\rm RMSNorm}(U U^{\top} x^{(i-1)})) + x^{(i-1)}, \\
            & = {\rm Bloc}k^{(i)}(U \cdot {\rm RMSNorm}(U^{\top} x^{(i-1)})) + x^{(i-1)}.
\end{align}    
\end{small}

\noindent Suppose $\hat{x}^{(i - 1)} = U^{\top} x^{(i-1)}$, $\hat{x}^{(i)} = U^{'\top} x^{(i)}$. Then, we have

\begin{small}
\begin{align}
    \hat{x}^{(i)} & = U^{'\top} {\rm Bloc}k^{(i)}(U \cdot {\rm RMSNorm}(\hat{x}^{(i-1)})) +  U^{'\top} U \hat{x}^{(i-1)}.
\end{align}    
\end{small}

\noindent It can be seen that if we make $U'$ equal to $U$, we don't need to add extra matrices at the residuals, because at this point we have $U^{'\top}U = I$. We refer to this unified computation of the main components of features in several consecutive blocks as Group PCA Projection. For LLM, the matrix introduced when the group size is 8 only increases the number of model parameters by 1\%.

It is also worth noting that Group PCA Projection requires models to use RMSNorm with a pre-norm architecture, but fortunately, based on existing work \cite{pre-rmsnorm.jiang2023}, all models that use LayerNorm can be directly converted to use RMSNorm. Moreover, nearly all LLMs use the pre-norm architecture.

\subsection{Experiments}

\vpara{Setup.} Following existing works on pruning LLM \cite{shearedllm.xia2023, llm-pruner.ma2023}. we consider the commonsense reasoning and question-answering tasks including Hellaswag \cite{hellaswag.zellers2019}, OpenBookQA \cite{OpenBookQA2018}, WinoGrande \cite{winogrande.sakaguchi2021}, ARC-Easy and ARC-Challenge \cite{ARC.clark2018}, BoolQ \cite{boolq.clark2019}, and PIQA \cite{piqa.bisk2020}. We evaluate our models using the Language Model Evaluation Harness \cite{eval-harness} framework. We use the SFT dataset oasst\_top1\_2023-08-25 to prune the TinyLlaMa-1.1B-Chat-v0.3 \cite{tinyllama.Zhang2023}. Without additional declarations, for TinyLlama we use Group PCA Projection with group size 8. 

% Meanwhile, it is worth noting that the parameters do not include the parameters introduced at the residuals.

\vpara{Results.} The experimental results on TinyLlaMa are shown in Table \ref{Tab.tinyllama}. We finetune the model on the dataset for $2$ epochs with a learning rate of 1e-5 and a batch size of 4 to obtain the pruned model. Then, we test the performance of the pruned model under a zero-shot setting. Our method updates both the model parameters and the mask. Since for limited SFT data, training too many epochs may lead to model overfitting, while too few updates can lead to non-convergence of the mask. Therefore we use smaller batch sizes so that convergence of the mask is guaranteed in case the model does not overfit the data.

% We use a very small batch size, this is because, with a limited amount of data, we need enough optimization steps to optimize all the masks.

\vpara{Impact of Group Size.} We also explore the effect of using different group sizes in Group PCA Projection on model performance. The experimental results are shown in Table \ref{Tab.tinyllama-abl-gs}. To prevent the interference of additional factors, we only consider the compression of the hidden dimension in this experiment. From the experimental results, it can be seen that using a smaller group size does not improve the performance of the model. On the contrary, a group size of 8 can be utilized without introducing too many additional parameters and performs better than a smaller group size. We speculate that larger group sizes with fewer additional parameters at residuals act as a form of regularization, helping to prevent overfitting of the calibration data by the results of principal component analysis.

\vpara{Combining with Existing Pruning Methods.} Our method can also be combined with other pruning methods, e.g., LLM-Pruner \cite{llm-pruner.ma2023}. For these methods, we first initialize the compactors using PCA projection, after which the compactors are directly fused to the model. Additionally, to maintain compatibility with existing pruning methods and avoid altering the model's structure, we utilize the same principal component matrix across all layers. In this way, $\rm SP^3$ can be viewed as a re-representation of the model making it easier to be pruned. In this experiment, we combine $\rm SP^3$ with LLM-Pruner. The experimental results are shown in Table \ref{Tab.tinyllama-llm-pruner}. It can be seen that combining our method can improve the performance of the LLM-pruner.

%% file: appendix/more_ablation.tex
\section{More Ablation Experiments}
\label{Appendix.more_ablation}

\subsection{Impact of Multi-level Masks}

% -----------------------------------------------------------------------------------------
\begin{table*}[t]
\newcolumntype{?}{!{\vrule width 1pt}}
\newcolumntype{C}{>{\centering\arraybackslash}p{2em}}
\centering
\small

\resizebox{0.5\linewidth}{!}{
\begin{tabular}{lcccc}
\toprule
           & SST-2 & RTE  & STS-B & MRPC \\ \midrule
\model-5M     & \textbf{91.4}  & \textbf{66.4} & 86.1  & 82.8 \\
\quad w/o head     & 90.9 & 66.1  & \textbf{86.6}  & 82.1     \\
\quad w/o layer    & 91.2 & 62.8  & 86.1  & 83.5     \\ 
\quad w/o head \& layer & 91.1 & 63.9 & 86.5  & \textbf{83.8}     \\ \bottomrule
\end{tabular}
}
\caption{Ablation studies of different levels of masks on datasets SST-2, RTE, STS-B, and MRPC using $\rm BERT_{base}$.}
\label{Tab.more_ablation}
\end{table*}
% -----------------------------------------------------------------------------------------

\smodel uses different levels of masks to compress the model. To explore the impact of different levels of masks on model performance, we conduct the following experiments: (1) Use all levels of masks. (2) Ignore head-level masks. (3) Ignore layer-level masks. (4) Ignore head-level and layer-level masks. 

The experimental results  are presented in Table \ref{Tab.more_ablation}. Observations indicate superior model performance on the SST-2 and RTE datasets when all mask levels are utilized. For the STS-B datasets, the removal of the head-level mask results in the most precise models. Notably, the optimal performance on the MRPC dataset is achieved by a model that excludes both the head-level and layer-level masks. Given the data volume in each dataset, we hypothesize that minor alterations in the head-level and layer-level masks can significantly influence model outputs compared to the dimension-level mask. This implies that the head-level and layer-level masks might be more challenging to optimize. Consequently, removing either the head-level or layer-level mask in smaller datasets can stabilize the optimization process, leading to a more precise model. As dataset sizes increase, the need for model compression flexibility becomes evident, with multi-level masking yielding superior outcomes.

\subsection{Impact of Different Random Seeds}

% -----------------------------------------------------------------------------------------

\begin{table*}[t]
\newcolumntype{?}{!{\vrule width 1pt}}
\newcolumntype{C}{>{\centering\arraybackslash}p{2em}}
\centering
\small

\resizebox{0.7\linewidth}{!}{
\begin{tabular}{c|c|cccccc|cc}
\toprule
          & Random seed      & 0    & 1    & 2    & 3    & 4    & 42   & Mean & Std  \\ \midrule
$\rm BERT_{base}$ & MPRC  & 82.8 & 81.3 & 81.8 & 83.8 & 83.1 & 82.8 & 82.6 & 0.82 \\
          & STS-B & 86.0 & 86.6 & 85.7 & 85.9 & 86.1 & 86.1 & 86.1 & 0.29 \\
          & RTE   & 64.2 & 62.5 & 62.8 & 64.2 & 63.2 & 66.1 & 63.8 & 1.2  \\ 
          & QNLI  & 86.9 & 86.9 & 87.3 & 87.1 & 86.9 & 87.6 & 87.1 & 0.26  \\ 
          & SST-2 & 91.1 & 91.3 & 91.3 & 90.8 & 90.7 & 91.4 & 91.1 & 0.26  \\ \midrule
$\rm OPT_{125m}$  & MPRC  & 78.9 & 78.4 & 78.1 & 76.9 & 76.7 & 77.2 & 77.7 & 0.81 \\
          & STS-B & 75.9 & 75.4 & 75.6 & 75.6 & 75.6 & 75.6 & 75.6 & 0.15 \\
          & RTE   & 57.7 & 58.6 & 57.8 & 59.6 & 57.5 & 60.3 & 58.6 & 1.0  \\ \bottomrule
\end{tabular}
}
\caption{Pruning performance  on datasets QNLI, SST-2, MRPC, STS-B, and RTE via the calibration data sampled by various random seeds.}
\label{Tab.more_ablation.random_seeds}
\end{table*}

% -----------------------------------------------------------------------------------------

$\rm SP^3$ needs to use the calibration data to initialize all the compactors. To verify the robustness of $\rm SP^3$ under different calibration data, we conduct experiments using different random seeds to sample the calibration data. We use $\rm SP^3$ to prune $\rm BERT_{base}$ and $\rm OPT_{125m}$ on datasets QNLI, SST-2, MRPC, STS-B, and RTE using a range of random seeds (0, 1, 2, 3, 4, 42), where $\rm BERT_{base}$ are pruned to 5M and $\rm OPT_{125m}$ are pruned to 7M. The results of the experiment are shown in Table \ref{Tab.more_ablation.random_seeds}. Employing these six random seeds has produced various calibration sets, and our experimental findings indicate that these different calibration sets have a minimal impact on the model's performance after compression due to their small standard deviation. 

%% file: appendix/structures.tex
\section{Structures of Pruned Model}
\label{Appendix.Structures}

The structure of pruned models on RTE, SST-2, STS-B, MNLI, QNLI, QQP and SQuAD are shown in Fig. \ref{Fig.Model.1} and Fig. \ref{Fig.Model.2}. We show the model structure in terms of dimension-level and head-level sparsity. Rather than directly displaying layer-level sparsity, we indirectly illustrate it through histograms depicting dimension-level and head-level sparsity. If a value at a certain position is 0, it suggests layer-level pruning has taken place at that position.

%% file: appendix/low_rank.tex
\section{Difference Between \smodel and Low-rank Factorization Methods}
\label{Appendix.comparison_with_low_rank}

Existing low-rank factorization methods \cite{cnn-pca.garg2019, tucker.liu2022, Tensorized.ma2019, dnnpca.riera2022, tucker.yin2022, CPTensor.zhou2019, pca-pruner.zhang2022} primarily focus on decomposing the weight matrix using PCA. In contrast, our approach applies PCA to the features generated during the model inference process and re-represents these features based on the principal components. This feature representation enhances compression efficacy and offers a distinct perspective from traditional weight matrix decomposition using PCA.

SliceGPT \cite{slicegpt.ashkboos2024} stands out as the most pertinent work to our research, as it focuses on model compression via principal component analysis of the feature matrix. However, notable distinctions exist between our approach and SliceGPT. Firstly, SliceGPT overlooks the variability in principal component distributions across layers, opting for a uniform compression rate of hidden dimension $d$ across all layers, while our approach adaptively determines the compression rate for different layers through learnable masks. Secondly, it neglects strategies for eliminating parameters introduced by residuals, which affects the compression rate of the compressed model. To address this problem, we instead propose the Group PCA Projection approach to reduce the number of parameters added at the residuals. Lastly, SliceGPT exclusively addresses decoder-only model architectures, omitting consideration for other model configurations. In contrast, we consider models with a decoder architecture, such as Llama, and OPT, as well as models with an encoder architecture, such as BERT.

% \subsection{Connection between \smodel and re-parameterization methods}

% Recently, re-parameterization methods have been proposed that can be used to improve existing model compression methods such as pruning~\cite{resrep.ding2021} and distillation~\cite{weight-distill.lin-etal-2021, WID.wu2023}. Its core idea is to represent the parameters of the compressed model based on the original model parameters. Therefore, this method can utilize the parameters of the original model more efficiently, which leads to better performance of the compressed model. \smodel can also be regarded as a re-parameterization method, and the major difference between it and the existing methods is that we construct it based on PCA theory. 

% Also, our approach suggests an explanation for the role of the re-parameterization approach, which is to preserve the main components of the features.

%% file: appendix/figures.tex
% -----------------------------------------------------------------------------------------
\newpage

\begin{figure*}[b] %H为当前位置，!htb为忽略美学标准，htbp为浮动图形

\subfigure[RTE: Hidden dimensions]{
\centering %图片居中
\includegraphics[width=0.3\textwidth]{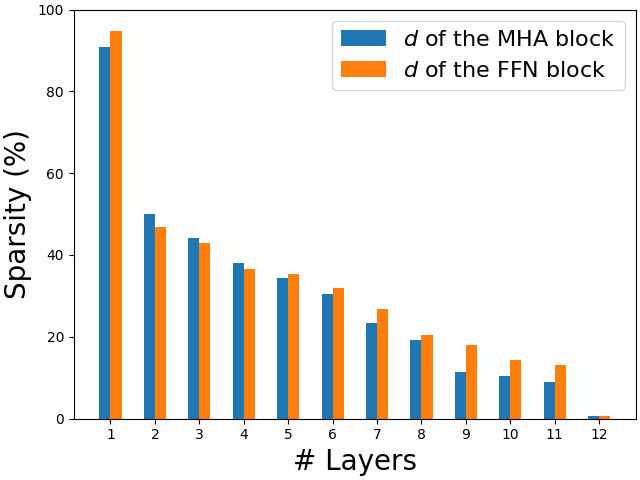} %插入图片，[]中设置图片大小，{}中是图片文件名
}
\subfigure[RTE: Intermediate dimensions]{
\centering %图片居中
\includegraphics[width=0.3\textwidth]{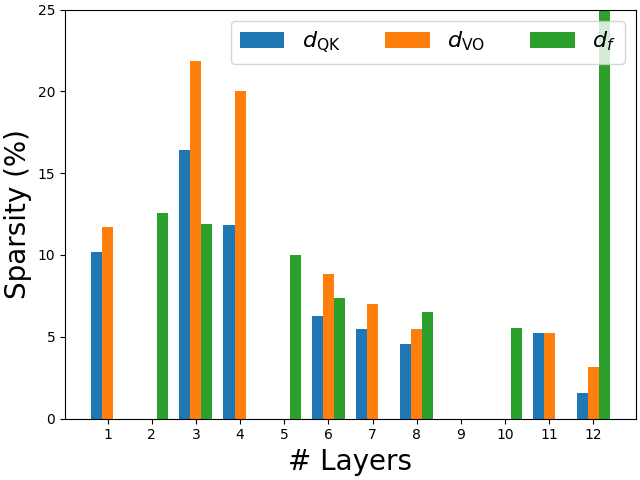} %插入图片，[]中设置图片大小，{}中是图片文件名
}
\subfigure[RTE: Attention heads size]{
\centering %图片居中
\includegraphics[width=0.3\textwidth]{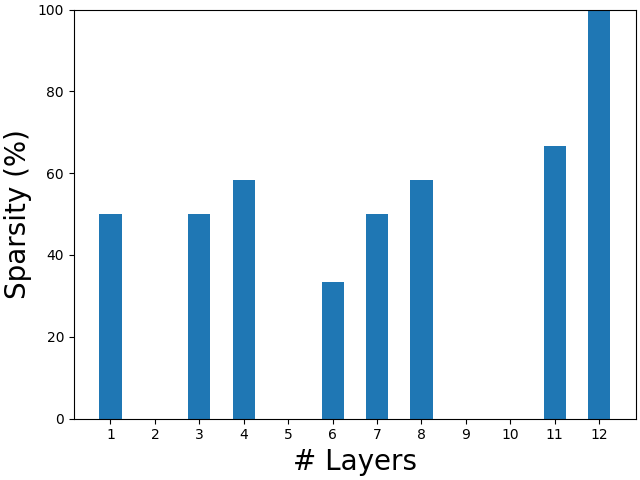} %插入图片，[]中设置图片大小，{}中是图片文件名
} \\

\subfigure[SST-2: Hidden dimensions]{
\centering %图片居中
\includegraphics[width=0.3\textwidth]{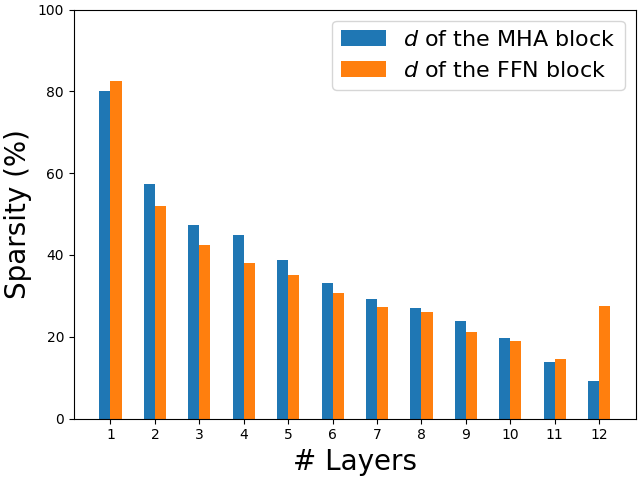} %插入图片，[]中设置图片大小，{}中是图片文件名
}
\subfigure[SST-2: Intermediate dimensions]{
\centering %图片居中
\includegraphics[width=0.3\textwidth]{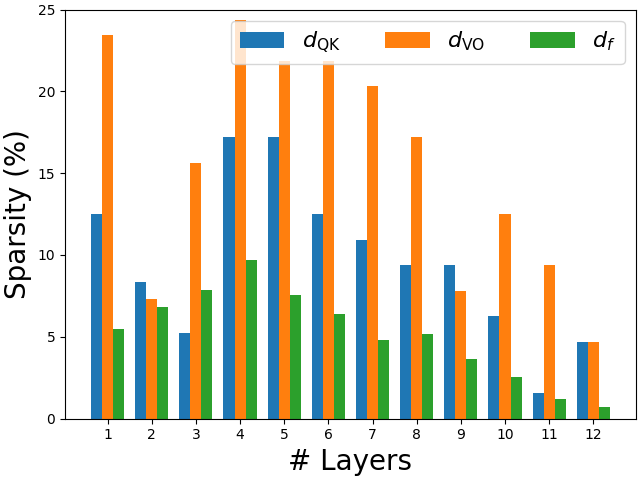} %插入图片，[]中设置图片大小，{}中是图片文件名
}
\subfigure[SST-2: Attention heads size]{
\centering %图片居中
\includegraphics[width=0.3\textwidth]{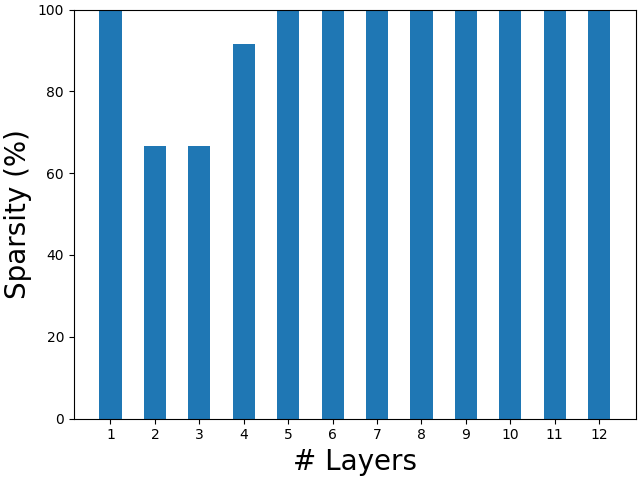} %插入图片，[]中设置图片大小，{}中是图片文件名
} \\

\subfigure[STS-B: Hidden dimensions]{
\centering %图片居中
\includegraphics[width=0.3\textwidth]{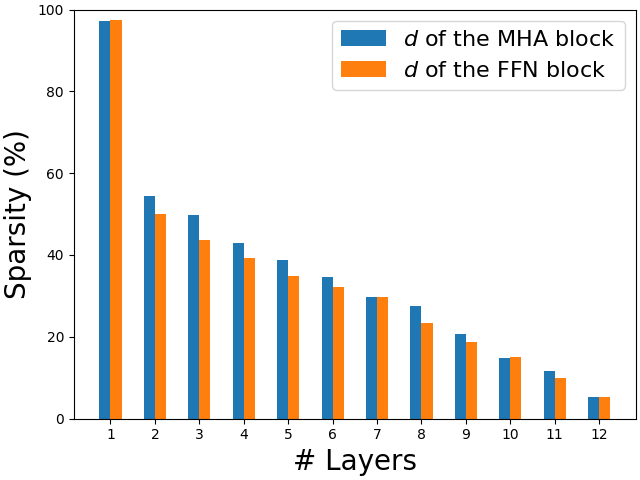} %插入图片，[]中设置图片大小，{}中是图片文件名
}
\subfigure[STS-B: Intermediate dimensions]{
\centering %图片居中
\includegraphics[width=0.3\textwidth]{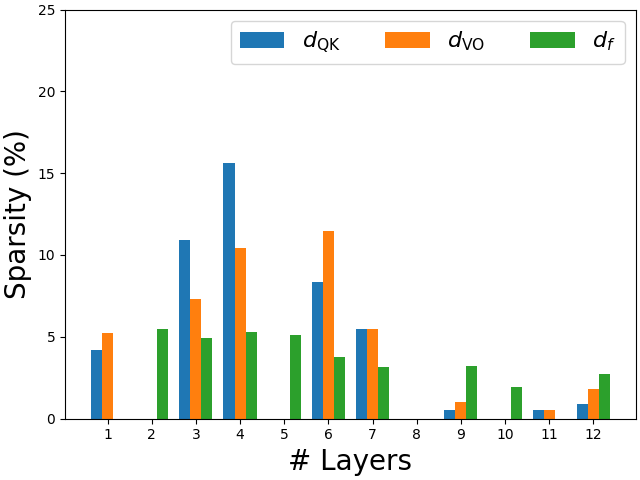} %插入图片，[]中设置图片大小，{}中是图片文件名
}
\subfigure[STS-B: Attention heads size]{
\centering %图片居中
\includegraphics[width=0.3\textwidth]{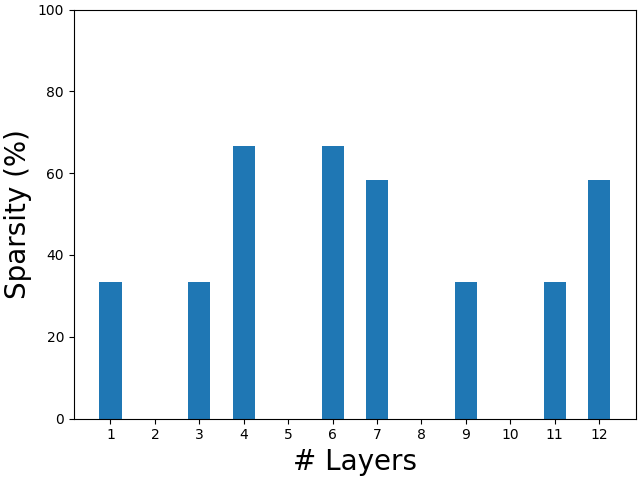} %插入图片，[]中设置图片大小，{}中是图片文件名
} 

\caption{\label{Fig.Model.1}Pruned model structures on RTE, SST-2 and STS-B datasets}

\end{figure*}

\begin{figure*}[b] %H为当前位置，!htb为忽略美学标准，htbp为浮动图形

\subfigure[MNLI: Hidden dimensions]{
\centering %图片居中
\includegraphics[width=0.3\textwidth]{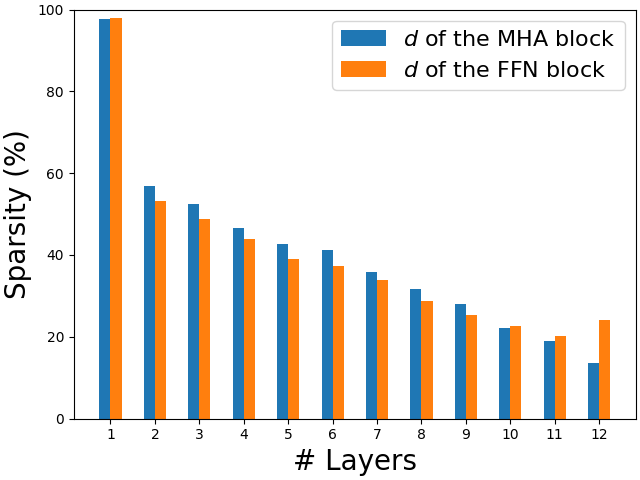} %插入图片，[]中设置图片大小，{}中是图片文件名
}
\subfigure[MNLI: Intermediate dimensions]{
\centering %图片居中
\includegraphics[width=0.3\textwidth]{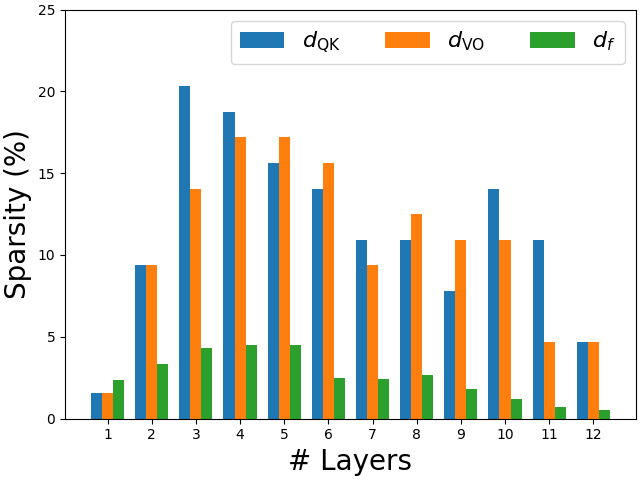} %插入图片，[]中设置图片大小，{}中是图片文件名
}
\subfigure[MNLI: Attention heads size]{
\centering %图片居中
\includegraphics[width=0.3\textwidth]{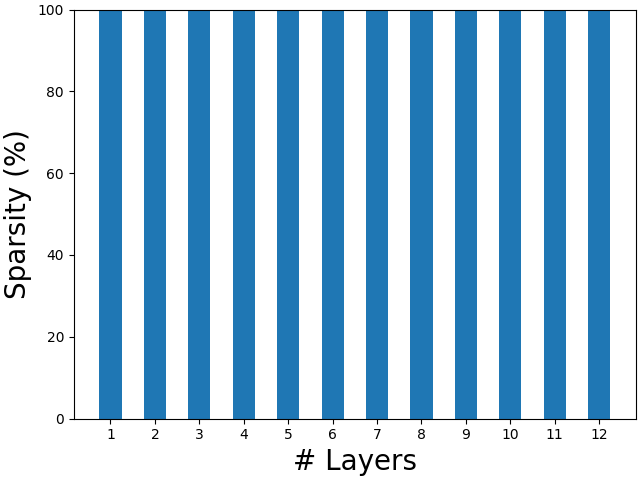} %插入图片，[]中设置图片大小，{}中是图片文件名
} \\

\subfigure[QNLI: Hidden dimensions]{
\centering %图片居中
\includegraphics[width=0.3\textwidth]{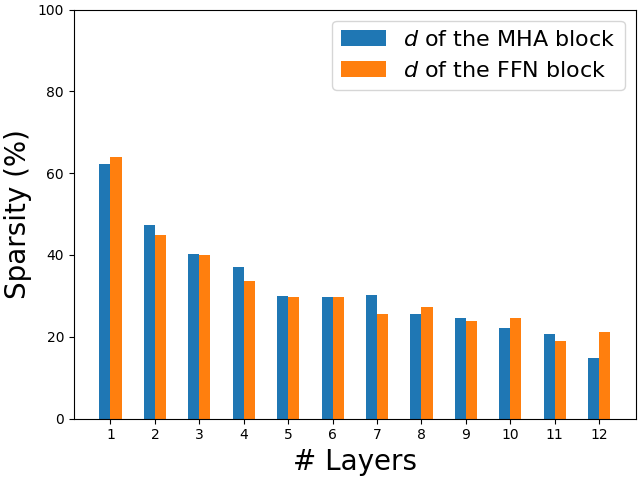} %插入图片，[]中设置图片大小，{}中是图片文件名
}
\subfigure[QNLI: Intermediate dimensions]{
\centering %图片居中
\includegraphics[width=0.3\textwidth]{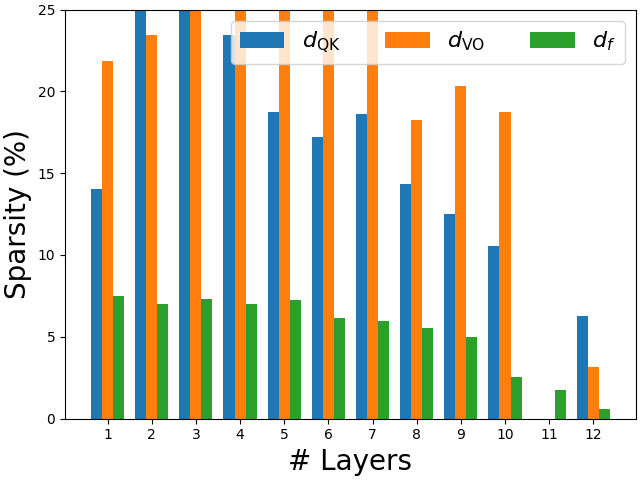} %插入图片，[]中设置图片大小，{}中是图片文件名
}
\subfigure[QNLI: Attention heads size]{
\centering %图片居中
\includegraphics[width=0.3\textwidth]{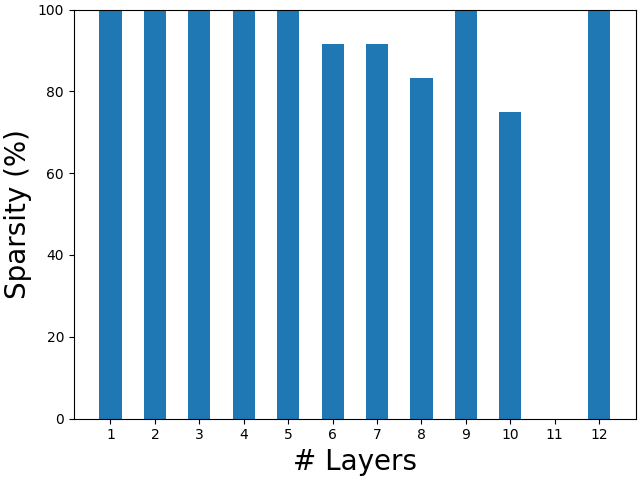} %插入图片，[]中设置图片大小，{}中是图片文件名
} \\

\subfigure[QQP: Hidden dimensions]{
\centering %图片居中
\includegraphics[width=0.3\textwidth]{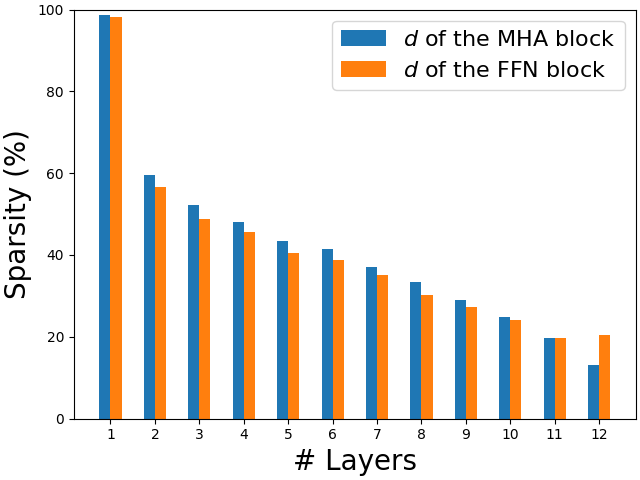} %插入图片，[]中设置图片大小，{}中是图片文件名
}
\subfigure[QQP: Intermediate dimensions]{
\centering %图片居中
\includegraphics[width=0.3\textwidth]{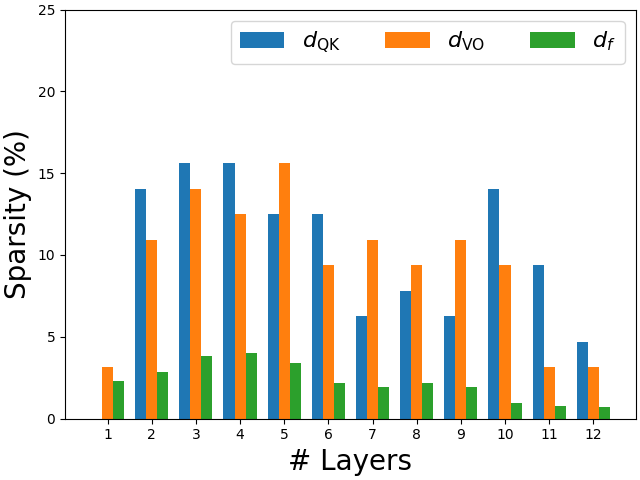} %插入图片，[]中设置图片大小，{}中是图片文件名
}
\subfigure[QQP: Attention heads size]{
\centering %图片居中
\includegraphics[width=0.3\textwidth]{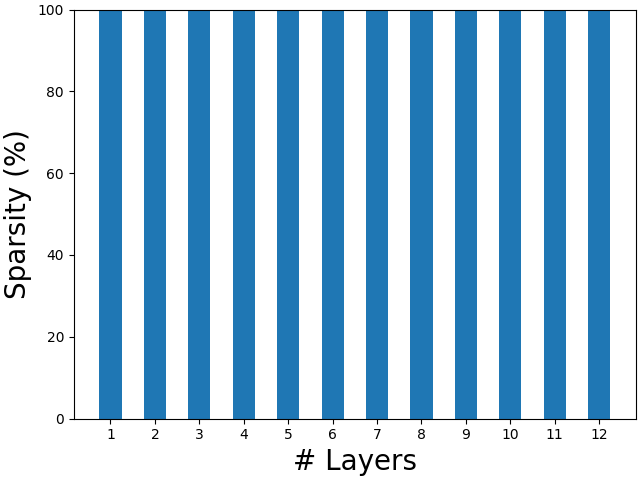} %插入图片，[]中设置图片大小，{}中是图片文件名
} \\

\subfigure[SQuAD: Hidden dimensions]{
\centering %图片居中
\includegraphics[width=0.3\textwidth]{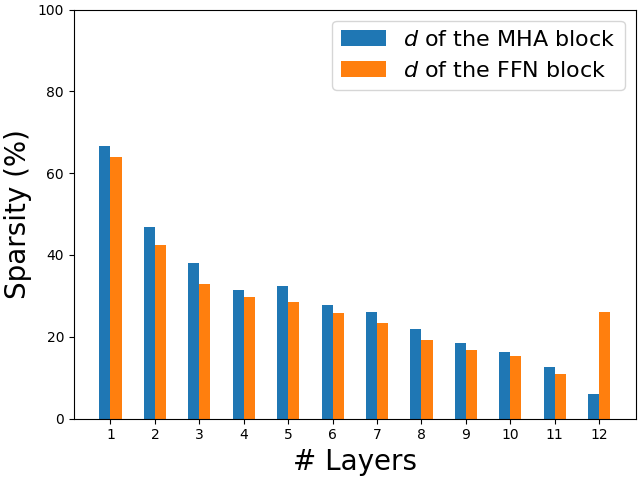} %插入图片，[]中设置图片大小，{}中是图片文件名
}
\subfigure[SQuAD: Intermediate dimensions]{
\centering %图片居中
\includegraphics[width=0.3\textwidth]{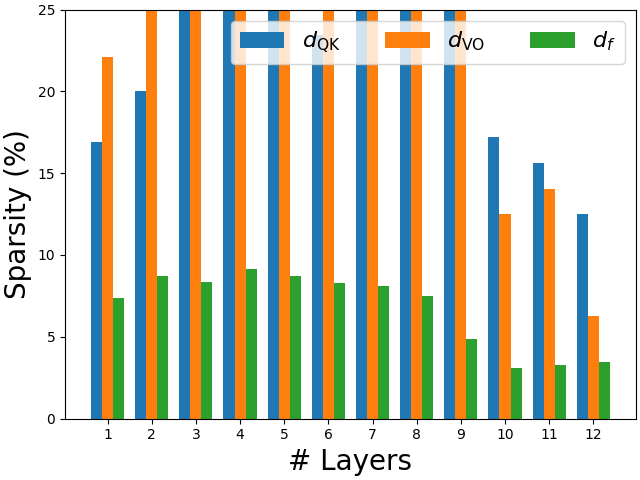} %插入图片，[]中设置图片大小，{}中是图片文件名
}
\subfigure[SQuAD: Attention heads size]{
\centering %图片居中
\includegraphics[width=0.3\textwidth]{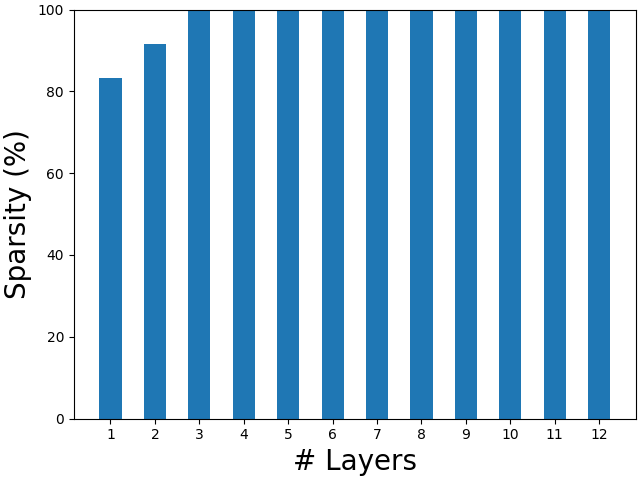} %插入图片，[]中设置图片大小，{}中是图片文件名
} 

\caption{\label{Fig.Model.2}Pruned model structures on MNLI, QNLI, QQP and SQuAD datasets}

\end{figure*}

% -----------------------------------------------------------------------------------------

%% file: main.bbl
\begin{thebibliography}{71}
\expandafter\ifx\csname natexlab\endcsname\relax\def\natexlab#1{#1}\fi

\bibitem[{Ashkboos et~al.(2024)Ashkboos, Croci, Nascimento, Hoefler, and Hensman}]{slicegpt.ashkboos2024}
Saleh Ashkboos, Maximilian~L Croci, Marcelo Gennari~do Nascimento, Torsten Hoefler, and James Hensman. 2024.
\newblock Slicegpt: Compress large language models by deleting rows and columns.
\newblock \emph{arXiv preprint arXiv:2401.15024}.

\bibitem[{Ba et~al.(2016)Ba, Kiros, and Hinton}]{Layernorm.Jimmy2016}
Jimmy~Lei Ba, Jamie~Ryan Kiros, and Geoffrey~E. Hinton. 2016.
\newblock \href {http://arxiv.org/abs/1607.06450} {Layer normalization}.

\bibitem[{Bisk et~al.(2020)Bisk, Zellers, Gao, Choi et~al.}]{piqa.bisk2020}
Yonatan Bisk, Rowan Zellers, Jianfeng Gao, Yejin Choi, et~al. 2020.
\newblock Piqa: Reasoning about physical commonsense in natural language.
\newblock In \emph{Proceedings of the AAAI conference on artificial intelligence}, volume~34, pages 7432--7439.

\bibitem[{Chen et~al.(2021)Chen, Yu, Dhillon, and Hsieh}]{Drone.chen2021}
Patrick Chen, Hsiang-Fu Yu, Inderjit Dhillon, and Cho-Jui Hsieh. 2021.
\newblock Drone: Data-aware low-rank compression for large nlp models.
\newblock \emph{Advances in neural information processing systems}, 34:29321--29334.

\bibitem[{Clark et~al.(2019)Clark, Lee, Chang, Kwiatkowski, Collins, and Toutanova}]{boolq.clark2019}
Christopher Clark, Kenton Lee, Ming-Wei Chang, Tom Kwiatkowski, Michael Collins, and Kristina Toutanova. 2019.
\newblock Boolq: Exploring the surprising difficulty of natural yes/no questions.
\newblock In \emph{Proceedings of the 2019 Conference of the North American Chapter of the Association for Computational Linguistics: Human Language Technologies, Volume 1 (Long and Short Papers)}, pages 2924--2936.

\bibitem[{Clark et~al.(2018)Clark, Cowhey, Etzioni, Khot, Sabharwal, Schoenick, and Tafjord}]{ARC.clark2018}
Peter Clark, Isaac Cowhey, Oren Etzioni, Tushar Khot, Ashish Sabharwal, Carissa Schoenick, and Oyvind Tafjord. 2018.
\newblock Think you have solved question answering? try arc, the ai2 reasoning challenge.
\newblock \emph{arXiv preprint arXiv:1803.05457}.

\bibitem[{Devlin et~al.(2018)Devlin, Chang, Lee, and Toutanova}]{BERT.devlin2018}
Jacob Devlin, Ming-Wei Chang, Kenton Lee, and Kristina Toutanova. 2018.
\newblock Bert: Pre-training of deep bidirectional transformers for language understanding.
\newblock \emph{arXiv preprint arXiv:1810.04805}.

\bibitem[{Ding et~al.(2021)Ding, Hao, Tan, Liu, Han, Guo, and Ding}]{resrep.ding2021}
Xiaohan Ding, Tianxiang Hao, Jianchao Tan, Ji~Liu, Jungong Han, Yuchen Guo, and Guiguang Ding. 2021.
\newblock Resrep: Lossless cnn pruning via decoupling remembering and forgetting.
\newblock In \emph{Proceedings of the IEEE/CVF International Conference on Computer Vision}, pages 4510--4520.

\bibitem[{Dolan and Brockett(2005)}]{mrpc.dolan2005}
William~B. Dolan and Chris Brockett. 2005.
\newblock \href {https://aclanthology.org/I05-5002} {Automatically constructing a corpus of sentential paraphrases}.
\newblock In \emph{Proceedings of the Third International Workshop on Paraphrasing ({IWP}2005)}.

\bibitem[{Fan et~al.(2019)Fan, Grave, and Joulin}]{Reducing.fan2019}
Angela Fan, Edouard Grave, and Armand Joulin. 2019.
\newblock Reducing transformer depth on demand with structured dropout.
\newblock In \emph{International Conference on Learning Representations}.

\bibitem[{Frankle and Carbin(2018)}]{lottery.jonathan2018}
Jonathan Frankle and Michael Carbin. 2018.
\newblock \href {http://arxiv.org/abs/1803.03635} {The lottery ticket hypothesis: Training pruned neural networks}.
\newblock \emph{CoRR}, abs/1803.03635.

\bibitem[{Fu et~al.(2021)Fu, Zhou, Yang, Tang, Liu, Liu, and Li}]{lrc-bert.fu2021}
Hao Fu, Shaojun Zhou, Qihong Yang, Junjie Tang, Guiquan Liu, Kaikui Liu, and Xiaolong Li. 2021.
\newblock Lrc-bert: latent-representation contrastive knowledge distillation for natural language understanding.
\newblock In \emph{Proceedings of the AAAI Conference on Artificial Intelligence}, volume~35, pages 12830--12838.

\bibitem[{Gale et~al.(2019)Gale, Elsen, and Hooker}]{State.gale2019}
Trevor Gale, Erich Elsen, and Sara Hooker. 2019.
\newblock The state of sparsity in deep neural networks.
\newblock \emph{arXiv preprint arXiv:1902.09574}.

\bibitem[{Gao et~al.(2023)Gao, Tow, Abbasi, Biderman, Black, DiPofi, Foster, Golding, Hsu, Le~Noac'h, Li, McDonell, Muennighoff, Ociepa, Phang, Reynolds, Schoelkopf, Skowron, Sutawika, Tang, Thite, Wang, Wang, and Zou}]{eval-harness}
Leo Gao, Jonathan Tow, Baber Abbasi, Stella Biderman, Sid Black, Anthony DiPofi, Charles Foster, Laurence Golding, Jeffrey Hsu, Alain Le~Noac'h, Haonan Li, Kyle McDonell, Niklas Muennighoff, Chris Ociepa, Jason Phang, Laria Reynolds, Hailey Schoelkopf, Aviya Skowron, Lintang Sutawika, Eric Tang, Anish Thite, Ben Wang, Kevin Wang, and Andy Zou. 2023.
\newblock \href {https://doi.org/10.5281/zenodo.10256836} {A framework for few-shot language model evaluation}.

\bibitem[{Garg et~al.(2019)Garg, Panda, and Roy}]{cnn-pca.garg2019}
Isha Garg, Priyadarshini Panda, and Kaushik Roy. 2019.
\newblock A low effort approach to structured cnn design using pca.
\newblock \emph{IEEE Access}, 8:1347--1360.

\bibitem[{Hinton et~al.(2015)Hinton, Vinyals, and Dean}]{distilling.hinton2015}
Geoffrey Hinton, Oriol Vinyals, and Jeff Dean. 2015.
\newblock Distilling the knowledge in a neural network.
\newblock \emph{arXiv preprint arXiv:1503.02531}.

\bibitem[{Hou et~al.(2020)Hou, Huang, Shang, Jiang, Chen, and Liu}]{Dynabert.hou2020}
Lu~Hou, Zhiqi Huang, Lifeng Shang, Xin Jiang, Xiao Chen, and Qun Liu. 2020.
\newblock Dynabert: Dynamic bert with adaptive width and depth.
\newblock \emph{Advances in Neural Information Processing Systems}, 33:9782--9793.

\bibitem[{Hua et~al.(2022)Hua, Hsu, Wang, Lou, Shen, and Jin}]{TFWSVD.hua2022}
Ting Hua, Yen-Chang Hsu, Felicity Wang, Qian Lou, Yilin Shen, and Hongxia Jin. 2022.
\newblock \href {https://aclanthology.org/2022.emnlp-main.91} {Numerical optimizations for weighted low-rank estimation on language models}.
\newblock In \emph{Proceedings of the 2022 Conference on Empirical Methods in Natural Language Processing}, pages 1404--1416, Abu Dhabi, United Arab Emirates. Association for Computational Linguistics.

\bibitem[{Jiang et~al.(2023)Jiang, Gu, Zhu, and Pan}]{pre-rmsnorm.jiang2023}
Zixuan Jiang, Jiaqi Gu, Hanqing Zhu, and David~Z Pan. 2023.
\newblock Pre-rmsnorm and pre-crmsnorm transformers: Equivalent and efficient pre-ln transformers.
\newblock \emph{arXiv preprint arXiv:2305.14858}.

\bibitem[{Jiao et~al.(2020)Jiao, Yin, Shang, Jiang, Chen, Li, Wang, and Liu}]{TinyBERT.jiao2020}
Xiaoqi Jiao, Yichun Yin, Lifeng Shang, Xin Jiang, Xiao Chen, Linlin Li, Fang Wang, and Qun Liu. 2020.
\newblock Tinybert: Distilling bert for natural language understanding.
\newblock In \emph{Findings of the Association for Computational Linguistics: EMNLP 2020}, pages 4163--4174.

\bibitem[{Kim et~al.(2019)Kim, Kang, and Kwak}]{mnli.kim2019}
Seonhoon Kim, Inho Kang, and Nojun Kwak. 2019.
\newblock Semantic sentence matching with densely-connected recurrent and co-attentive information.
\newblock In \emph{Proceedings of the AAAI conference on artificial intelligence}, volume~33, pages 6586--6593.

\bibitem[{Kurtic et~al.(2022)Kurtic, Campos, Nguyen, Frantar, Kurtz, Fineran, Goin, and Alistarh}]{Optimal.kurtic2022}
Eldar Kurtic, Daniel Campos, Tuan Nguyen, Elias Frantar, Mark Kurtz, Benjamin Fineran, Michael Goin, and Dan Alistarh. 2022.
\newblock \href {https://aclanthology.org/2022.emnlp-main.279} {The optimal {BERT} surgeon: Scalable and accurate second-order pruning for large language models}.
\newblock In \emph{Proceedings of the 2022 Conference on Empirical Methods in Natural Language Processing}, pages 4163--4181, Abu Dhabi, United Arab Emirates. Association for Computational Linguistics.

\bibitem[{Kwon et~al.(2022)Kwon, Kim, Mahoney, Hassoun, Keutzer, and Gholami}]{Fast.kwon2022}
Woosuk Kwon, Sehoon Kim, Michael~W Mahoney, Joseph Hassoun, Kurt Keutzer, and Amir Gholami. 2022.
\newblock A fast post-training pruning framework for transformers.
\newblock \emph{arXiv preprint arXiv:2204.09656}.

\bibitem[{Lagunas et~al.(2021)Lagunas, Charlaix, Sanh, and Rush}]{BMP.lagunas2021}
Fran{\c{c}}ois Lagunas, Ella Charlaix, Victor Sanh, and Alexander Rush. 2021.
\newblock \href {https://doi.org/10.18653/v1/2021.emnlp-main.829} {Block pruning for faster transformers}.
\newblock In \emph{Proceedings of the 2021 Conference on Empirical Methods in Natural Language Processing}, pages 10619--10629, Online and Punta Cana, Dominican Republic. Association for Computational Linguistics.

\bibitem[{Li et~al.(2020)Li, Liu, Zhao, Xu, Yang, and Jin}]{bert-emd.li2020}
Jianquan Li, Xiaokang Liu, Honghong Zhao, Ruifeng Xu, Min Yang, and Yaohong Jin. 2020.
\newblock Bert-emd: Many-to-many layer mapping for bert compression with earth mover’s distance.
\newblock In \emph{Proceedings of the 2020 Conference on Empirical Methods in Natural Language Processing (EMNLP)}, pages 3009--3018.

\bibitem[{Lin et~al.(2021)Lin, Li, Wang, Li, Du, Xiao, and Zhu}]{weight-distill.lin-etal-2021}
Ye~Lin, Yanyang Li, Ziyang Wang, Bei Li, Quan Du, Tong Xiao, and Jingbo Zhu. 2021.
\newblock \href {https://doi.org/10.18653/v1/2021.acl-long.162} {Weight distillation: Transferring the knowledge in neural network parameters}.
\newblock In \emph{Proceedings of the 59th Annual Meeting of the Association for Computational Linguistics and the 11th International Joint Conference on Natural Language Processing (Volume 1: Long Papers)}, pages 2076--2088, Online. Association for Computational Linguistics.

\bibitem[{Lin et~al.(2020)Lin, Liu, Yang, Hua, and Roth}]{Slip.lin2020}
Zi~Lin, Jeremiah Liu, Zi~Yang, Nan Hua, and Dan Roth. 2020.
\newblock \href {https://doi.org/10.18653/v1/2020.findings-emnlp.64} {Pruning redundant mappings in transformer models via spectral-normalized identity prior}.
\newblock In \emph{Findings of the Association for Computational Linguistics: EMNLP 2020}, pages 719--730, Online. Association for Computational Linguistics.

\bibitem[{Liu and Ng(2022)}]{tucker.liu2022}
Ye~Liu and Michael~K Ng. 2022.
\newblock Deep neural network compression by tucker decomposition with nonlinear response.
\newblock \emph{Knowledge-Based Systems}, page 108171.

\bibitem[{Liu et~al.(2021)Liu, Lin, and Yuan}]{rosita.liu2021}
Yuanxin Liu, Zheng Lin, and Fengcheng Yuan. 2021.
\newblock Rosita: Refined bert compression with integrated techniques.
\newblock In \emph{Proceedings of the AAAI Conference on Artificial Intelligence}, volume~35, pages 8715--8722.

\bibitem[{Louizos et~al.(2018)Louizos, Welling, and Kingma}]{l0pruning.louizos2018}
Christos Louizos, Max Welling, and Diederik~P Kingma. 2018.
\newblock Learning sparse neural networks through l\_0 regularization.
\newblock In \emph{International Conference on Learning Representations}.

\bibitem[{Ma et~al.(2019)Ma, Zhang, Zhang, Duan, Hou, Zhou, and Song}]{Tensorized.ma2019}
Xindian Ma, Peng Zhang, Shuai Zhang, Nan Duan, Yuexian Hou, Ming Zhou, and Dawei Song. 2019.
\newblock A tensorized transformer for language modeling.
\newblock \emph{Advances in neural information processing systems}, 32.

\bibitem[{Ma et~al.(2023)Ma, Fang, and Wang}]{llm-pruner.ma2023}
Xinyin Ma, Gongfan Fang, and Xinchao Wang. 2023.
\newblock Llm-pruner: On the structural pruning of large language models.
\newblock \emph{arXiv preprint arXiv:2305.11627}.

\bibitem[{Ma{\'c}kiewicz and Ratajczak(1993)}]{PCA}
Andrzej Ma{\'c}kiewicz and Waldemar Ratajczak. 1993.
\newblock Principal components analysis (pca).
\newblock \emph{Computers \& Geosciences}, 19(3):303--342.

\bibitem[{McCarley et~al.(2019)McCarley, Chakravarti, and Sil}]{FFNPruning.mccarley2019}
JS~McCarley, Rishav Chakravarti, and Avirup Sil. 2019.
\newblock Structured pruning of a bert-based question answering model.
\newblock \emph{arXiv preprint arXiv:1910.06360}.

\bibitem[{Michel et~al.(2019)Michel, Levy, and Neubig}]{HeadPruning.michel2019}
Paul Michel, Omer Levy, and Graham Neubig. 2019.
\newblock Are sixteen heads really better than one?
\newblock \emph{Advances in neural information processing systems}, 32.

\bibitem[{Mihaylov et~al.(2018)Mihaylov, Clark, Khot, and Sabharwal}]{OpenBookQA2018}
Todor Mihaylov, Peter Clark, Tushar Khot, and Ashish Sabharwal. 2018.
\newblock Can a suit of armor conduct electricity? a new dataset for open book question answering.
\newblock In \emph{EMNLP}.

\bibitem[{Nova et~al.(2023)Nova, Dai, and Schuurmans}]{KCM.nova2023}
Azade Nova, Hanjun Dai, and Dale Schuurmans. 2023.
\newblock Gradient-free structured pruning with unlabeled data.
\newblock \emph{arXiv preprint arXiv:2303.04185}.

\bibitem[{Ouyang et~al.(2022)Ouyang, Wu, Jiang, Almeida, Wainwright, Mishkin, Zhang, Agarwal, Slama, Ray et~al.}]{instruct-gpt.ouyang2022}
Long Ouyang, Jeffrey Wu, Xu~Jiang, Diogo Almeida, Carroll Wainwright, Pamela Mishkin, Chong Zhang, Sandhini Agarwal, Katarina Slama, Alex Ray, et~al. 2022.
\newblock Training language models to follow instructions with human feedback.
\newblock \emph{Advances in Neural Information Processing Systems}, 35:27730--27744.

\bibitem[{Paszke et~al.(2019)Paszke, Gross, Massa, Lerer, Bradbury, Chanan, Killeen, Lin, Gimelshein, Antiga, Desmaison, K{\"{o}}pf, Yang, DeVito, Raison, Tejani, Chilamkurthy, Steiner, Fang, Bai, and Chintala}]{pytorch}
Adam Paszke, Sam Gross, Francisco Massa, Adam Lerer, James Bradbury, Gregory Chanan, Trevor Killeen, Zeming Lin, Natalia Gimelshein, Luca Antiga, Alban Desmaison, Andreas K{\"{o}}pf, Edward~Z. Yang, Zach DeVito, Martin Raison, Alykhan Tejani, Sasank Chilamkurthy, Benoit Steiner, Lu~Fang, Junjie Bai, and Soumith Chintala. 2019.
\newblock \href {http://arxiv.org/abs/1912.01703} {Pytorch: An imperative style, high-performance deep learning library}.
\newblock \emph{CoRR}, abs/1912.01703.

\bibitem[{Peiyuan~Zhang and Lu(2023)}]{tinyllama.Zhang2023}
Tianduo~Wang Peiyuan~Zhang, Guangtao~Zeng and Wei Lu. 2023.
\newblock \href {https://github.com/jzhang38/TinyLlama} {Tinyllama}.

\bibitem[{Prasanna et~al.(2020)Prasanna, Rogers, and Rumshisky}]{FFNPruning.prasanna2020}
Sai Prasanna, Anna Rogers, and Anna Rumshisky. 2020.
\newblock When bert plays the lottery, all tickets are winning.
\newblock \emph{arXiv preprint arXiv:2005.00561}.

\bibitem[{Radford et~al.()Radford, Wu, Child, Luan, Amodei, Sutskever et~al.}]{GPT2.radford2019}
Alec Radford, Jeffrey Wu, Rewon Child, David Luan, Dario Amodei, Ilya Sutskever, et~al.
\newblock Language models are unsupervised multitask learners.

\bibitem[{Rajpurkar et~al.(2016)Rajpurkar, Zhang, Lopyrev, and Liang}]{SQuAD.2016}
Pranav Rajpurkar, Jian Zhang, Konstantin Lopyrev, and Percy Liang. 2016.
\newblock \href {https://doi.org/10.18653/v1/D16-1264} {{SQ}u{AD}: 100,000+ questions for machine comprehension of text}.
\newblock In \emph{Proceedings of the 2016 Conference on Empirical Methods in Natural Language Processing}, pages 2383--2392, Austin, Texas. Association for Computational Linguistics.

\bibitem[{Ren and Zhu(2023)}]{LPAF.ren2023}
Siyu Ren and Kenny~Q Zhu. 2023.
\newblock Low-rank prune-and-factorize for language model compression.
\newblock \emph{arXiv preprint arXiv:2306.14152}.

\bibitem[{Riera et~al.(2022)Riera, Arnau, and Gonz{\'a}lez}]{dnnpca.riera2022}
Marc Riera, Jose~Maria Arnau, and Antonio Gonz{\'a}lez. 2022.
\newblock Dnn pruning with principal component analysis and connection importance estimation.
\newblock \emph{Journal of Systems Architecture}, 122:102336.

\bibitem[{Sajjad et~al.(2020)Sajjad, Dalvi, Durrani, and Nakov}]{Poor.sajjad2020}
Hassan Sajjad, Fahim Dalvi, Nadir Durrani, and Preslav Nakov. 2020.
\newblock Poor man’s bert: Smaller and faster transformer models.
\newblock \emph{arXiv preprint arXiv:2004.03844}, 2(2).

\bibitem[{Sajjad et~al.(2023)Sajjad, Dalvi, Durrani, and Nakov}]{DropLayer.sajjad2023}
Hassan Sajjad, Fahim Dalvi, Nadir Durrani, and Preslav Nakov. 2023.
\newblock On the effect of dropping layers of pre-trained transformer models.
\newblock \emph{Computer Speech \& Language}, 77:101429.

\bibitem[{Sakaguchi et~al.(2021)Sakaguchi, Bras, Bhagavatula, and Choi}]{winogrande.sakaguchi2021}
Keisuke Sakaguchi, Ronan~Le Bras, Chandra Bhagavatula, and Yejin Choi. 2021.
\newblock Winogrande: An adversarial winograd schema challenge at scale.
\newblock \emph{Communications of the ACM}, 64(9):99--106.

\bibitem[{Sanh et~al.(2019)Sanh, Debut, Chaumond, and Wolf}]{Distillbert.VictorSanh}
Victor Sanh, Lysandre Debut, Julien Chaumond, and Thomas Wolf. 2019.
\newblock \href {http://arxiv.org/abs/1910.01108} {Distilbert, a distilled version of {BERT:} smaller, faster, cheaper and lighter}.
\newblock \emph{CoRR}, abs/1910.01108.

\bibitem[{Sanh et~al.(2020)Sanh, Wolf, and Rush}]{Movement.sanh2020}
Victor Sanh, Thomas Wolf, and Alexander Rush. 2020.
\newblock Movement pruning: Adaptive sparsity by fine-tuning.
\newblock \emph{Advances in Neural Information Processing Systems}, 33:20378--20389.

\bibitem[{Socher et~al.(2013)Socher, Perelygin, Wu, Chuang, Manning, Ng, and Potts}]{sst2.socher2013}
Richard Socher, Alex Perelygin, Jean Wu, Jason Chuang, Christopher~D Manning, Andrew~Y Ng, and Christopher Potts. 2013.
\newblock Recursive deep models for semantic compositionality over a sentiment treebank.
\newblock In \emph{Proceedings of the 2013 conference on empirical methods in natural language processing}, pages 1631--1642.

\bibitem[{Su et~al.(2024)Su, Ahmed, Lu, Pan, Bo, and Liu}]{rope.su2024}
Jianlin Su, Murtadha Ahmed, Yu~Lu, Shengfeng Pan, Wen Bo, and Yunfeng Liu. 2024.
\newblock Roformer: Enhanced transformer with rotary position embedding.
\newblock \emph{Neurocomputing}, 568:127063.

\bibitem[{Tahaei et~al.(2022)Tahaei, Charlaix, Nia, Ghodsi, and Rezagholizadeh}]{Kroneckerbert.tahaei2022}
Marzieh Tahaei, Ella Charlaix, Vahid Nia, Ali Ghodsi, and Mehdi Rezagholizadeh. 2022.
\newblock \href {https://doi.org/10.18653/v1/2022.naacl-main.154} {{K}ronecker{BERT}: Significant compression of pre-trained language models through kronecker decomposition and knowledge distillation}.
\newblock In \emph{Proceedings of the 2022 Conference of the North American Chapter of the Association for Computational Linguistics: Human Language Technologies}, pages 2116--2127, Seattle, United States. Association for Computational Linguistics.

\bibitem[{Touvron et~al.(2023)Touvron, Martin, Stone, Albert, Almahairi, Babaei, Bashlykov, Batra, Bhargava, Bhosale et~al.}]{llama.touvron2023}
Hugo Touvron, Louis Martin, Kevin Stone, Peter Albert, Amjad Almahairi, Yasmine Babaei, Nikolay Bashlykov, Soumya Batra, Prajjwal Bhargava, Shruti Bhosale, et~al. 2023.
\newblock Llama 2: Open foundation and fine-tuned chat models.
\newblock \emph{arXiv preprint arXiv:2307.09288}.

\bibitem[{Voita et~al.(2019)Voita, Talbot, Moiseev, Sennrich, and Titov}]{HeadPruning.voita2019}
Elena Voita, David Talbot, Fedor Moiseev, Rico Sennrich, and Ivan Titov. 2019.
\newblock Analyzing multi-head self-attention: Specialized heads do the heavy lifting, the rest can be pruned.
\newblock In \emph{Proceedings of the 57th Annual Meeting of the Association for Computational Linguistics}. Association for Computational Linguistics.

\bibitem[{Wang et~al.(2018)Wang, Singh, Michael, Hill, Levy, and Bowman}]{GLUE.wang2018}
Alex Wang, Amanpreet Singh, Julian Michael, Felix Hill, Omer Levy, and Samuel Bowman. 2018.
\newblock \href {https://doi.org/10.18653/v1/W18-5446} {{GLUE}: A multi-task benchmark and analysis platform for natural language understanding}.
\newblock In \emph{Proceedings of the 2018 {EMNLP} Workshop {B}lackbox{NLP}: Analyzing and Interpreting Neural Networks for {NLP}}, pages 353--355, Brussels, Belgium. Association for Computational Linguistics.

\bibitem[{Wang et~al.(2017)Wang, Hamza, and Florian}]{qqp.wang2017}
Zhiguo Wang, Wael Hamza, and Radu Florian. 2017.
\newblock Bilateral multi-perspective matching for natural language sentences.
\newblock \emph{arXiv preprint arXiv:1702.03814}.

\bibitem[{Wang et~al.(2020)Wang, Wohlwend, and Lei}]{Structured.wang2020}
Ziheng Wang, Jeremy Wohlwend, and Tao Lei. 2020.
\newblock Structured pruning of large language models.
\newblock In \emph{Proceedings of the 2020 Conference on Empirical Methods in Natural Language Processing (EMNLP)}, pages 6151--6162.

\bibitem[{Wolf et~al.(2019)Wolf, Debut, Sanh, Chaumond, Delangue, Moi, Cistac, Rault, Louf, Funtowicz, and Brew}]{huggingface}
Thomas Wolf, Lysandre Debut, Victor Sanh, Julien Chaumond, Clement Delangue, Anthony Moi, Pierric Cistac, Tim Rault, R{\'{e}}mi Louf, Morgan Funtowicz, and Jamie Brew. 2019.
\newblock \href {http://arxiv.org/abs/1910.03771} {Huggingface's transformers: State-of-the-art natural language processing}.
\newblock \emph{CoRR}, abs/1910.03771.

\bibitem[{Wu et~al.(2023{\natexlab{a}})Wu, Chen, Quan, Wang, and Wang}]{AD-KD.wu2023}
Siyue Wu, Hongzhan Chen, Xiaojun Quan, Qifan Wang, and Rui Wang. 2023{\natexlab{a}}.
\newblock Ad-kd: Attribution-driven knowledge distillation for language model compression.
\newblock \emph{arXiv preprint arXiv:2305.10010}.

\bibitem[{Wu et~al.(2023{\natexlab{b}})Wu, Hou, Zhao, Lao, Li, Wong, and Yang}]{WID.wu2023}
Taiqiang Wu, Cheng Hou, Zhe Zhao, Shanshan Lao, Jiayi Li, Ngai Wong, and Yujiu Yang. 2023{\natexlab{b}}.
\newblock Weight-inherited distillation for task-agnostic bert compression.
\newblock \emph{arXiv preprint arXiv:2305.09098}.

\bibitem[{Xia et~al.(2023)Xia, Gao, Zeng, and Chen}]{shearedllm.xia2023}
Mengzhou Xia, Tianyu Gao, Zhiyuan Zeng, and Danqi Chen. 2023.
\newblock Sheared llama: Accelerating language model pre-training via structured pruning.
\newblock \emph{arXiv preprint arXiv:2310.06694}.

\bibitem[{Xia et~al.(2022)Xia, Zhong, and Chen}]{cofi.xia2022}
Mengzhou Xia, Zexuan Zhong, and Danqi Chen. 2022.
\newblock Structured pruning learns compact and accurate models.
\newblock In \emph{Proceedings of the 60th Annual Meeting of the Association for Computational Linguistics (Volume 1: Long Papers)}, pages 1513--1528.

\bibitem[{Xiao et~al.(2023)Xiao, Yin, Gong, Zang, Ren, and Yuan}]{COMCAT.xiao2023}
Jinqi Xiao, Miao Yin, Yu~Gong, Xiao Zang, Jian Ren, and Bo~Yuan. 2023.
\newblock \href {https://proceedings.mlr.press/v202/xiao23e.html} {{COMCAT}: Towards efficient compression and customization of attention-based vision models}.
\newblock In \emph{Proceedings of the 40th International Conference on Machine Learning}, volume 202 of \emph{Proceedings of Machine Learning Research}, pages 38125--38136. PMLR.

\bibitem[{Xu et~al.(2020)Xu, Zhou, Ge, Wei, and Zhou}]{Bert-of-theseus.xu2020}
Canwen Xu, Wangchunshu Zhou, Tao Ge, Furu Wei, and Ming Zhou. 2020.
\newblock Bert-of-theseus: Compressing bert by progressive module replacing.
\newblock \emph{arXiv preprint arXiv:2002.02925}.

\bibitem[{Xu et~al.(2022)Xu, Luo, Wang, Chang, Huang, Huang, and Huang}]{CAP-f.xu2022}
Runxin Xu, Fuli Luo, Chengyu Wang, Baobao Chang, Jun Huang, Songfang Huang, and Fei Huang. 2022.
\newblock From dense to sparse: Contrastive pruning for better pre-trained language model compression.
\newblock In \emph{Proceedings of the AAAI Conference on Artificial Intelligence}, volume~36, pages 11547--11555.

\bibitem[{Yin et~al.(2022)Yin, Phan, Zang, Liao, and Yuan}]{tucker.yin2022}
Miao Yin, Huy Phan, Xiao Zang, Siyu Liao, and Bo~Yuan. 2022.
\newblock Batude: Budget-aware neural network compression based on tucker decomposition.
\newblock In \emph{Proceedings of the AAAI Conference on Artificial Intelligence}, 8, pages 8874--8882.

\bibitem[{Zellers et~al.(2019)Zellers, Holtzman, Bisk, Farhadi, and Choi}]{hellaswag.zellers2019}
Rowan Zellers, Ari Holtzman, Yonatan Bisk, Ali Farhadi, and Yejin Choi. 2019.
\newblock Hellaswag: Can a machine really finish your sentence?
\newblock In \emph{Proceedings of the 57th Annual Meeting of the Association for Computational Linguistics}, pages 4791--4800.

\bibitem[{Zhang et~al.(2022)Zhang, Zuo, Liang, Bukharin, He, Chen, and Zhao}]{platon.zhang2022}
Qingru Zhang, Simiao Zuo, Chen Liang, Alexander Bukharin, Pengcheng He, Weizhu Chen, and Tuo Zhao. 2022.
\newblock Platon: Pruning large transformer models with upper confidence bound of weight importance.
\newblock In \emph{International Conference on Machine Learning}, pages 26809--26823. PMLR.

\bibitem[{Zhang and Wang(2022)}]{pca-pruner.zhang2022}
Wei Zhang and Zhiming Wang. 2022.
\newblock Pca-pruner: Filter pruning by principal component analysis.
\newblock \emph{Journal of Intelligent \& Fuzzy Systems}, 43(4):4803--4813.

\bibitem[{Zhou et~al.(2019)Zhou, Liu, Long, Chen, and Zhu}]{CPTensor.zhou2019}
Mingyi Zhou, Yipeng Liu, Zhen Long, Longxi Chen, and Ce~Zhu. 2019.
\newblock Tensor rank learning in cp decomposition via convolutional neural network.
\newblock \emph{Signal Processing: Image Communication}, 73:12--21.

\end{thebibliography}
